\RequirePackage{fix-cm}
\documentclass[twocolumn]{svjour3}           

\makeatletter
\expandafter\let\csname opt@amsmath.sty\endcsname\relax 
\makeatother
\smartqed  

\usepackage{graphicx}
\usepackage{amsmath}
\usepackage[T1,hyphens]{url}
\usepackage[colorlinks,urlcolor=blue]{hyperref}
\usepackage{mathtools, cuted}
\usepackage{multirow}

\newcommand{\eg}{\emph{e.g.}}

\newcommand{\etal}{\emph{et al.}}
\newcommand{\ie}{\emph{i.e.}}
\DeclareMathOperator{\total}{\mathit{d}}

\begin{document}

\title{Cross-Domain Image Matching with Deep Feature Maps}


\author{Bailey Kong \and
        James Supan\u{c}i\u{c}, III \and
        Deva Ramanan \and
        Charless C. Fowlkes
}

\authorrunning{Kong et al.} 

\institute{B. Kong, J. Supancic, C. Fowlkes \at
              Department of Computer Science\\
              University of California\\
              Irvine, CA  92617\\
              \email{\{bhkong,jsupanci,fowlkes\}@ics.uci.edu}           
           \and
           D. Ramanan \at
           Robotics Institute\\
           Carnegie Mellon University\\
           Pittsburgh, PA  15213\\
           \email{deva@cs.cmu.edu}
}

\date{Received: date / Accepted: date}

\maketitle

\begin{abstract}
  We investigate the problem of automatically determining what type of shoe
  left an impression found at a crime scene. This recognition problem is made
  difficult by the variability in types of crime scene evidence (ranging from
  traces of dust or oil on hard surfaces to impressions made in soil) and the
  lack of comprehensive databases of shoe outsole tread patterns.  We find that
  mid-level features extracted by pre-trained convolutional neural nets are
  surprisingly effective descriptors for this specialized domains. However,
  the choice of similarity measure for matching exemplars to a query image is
  essential to good performance. For matching multi-channel deep features, we
  propose the use of \emph{multi-channel normalized cross-correlation} and
  analyze its effectiveness.  Our proposed metric significantly improves
  performance in matching crime scene shoeprints to laboratory test
  impressions. We also show its effectiveness in other cross-domain image
  retrieval problems: matching facade images to segmentation labels and aerial
  photos to map images.  Finally, we introduce a discriminatively trained
  variant and fine-tune our system through our proposed metric, obtaining
  state-of-the-art performance.
\end{abstract}

\section{Introduction}

We investigate the problem of automatically determining what type
(brand/model/size) of shoe left an impression found at a crime scene. In the
forensic footwear examination literature \cite{bodziak1999footwear}, this
fine-grained category-level recognition problem is known as determining the
\textit{class characteristics} of a tread impression.  This is distinct from
the instance-level recognition problem of matching \textit{acquired
characteristics} such as cuts or scratches which can provide stronger evidence
that a specific shoe left a specific mark.

\begin{figure*}[h]
\begin{center}
  \begin{tabular}{c}
  \includegraphics[trim=0 0 0 0,clip, width=160mm]{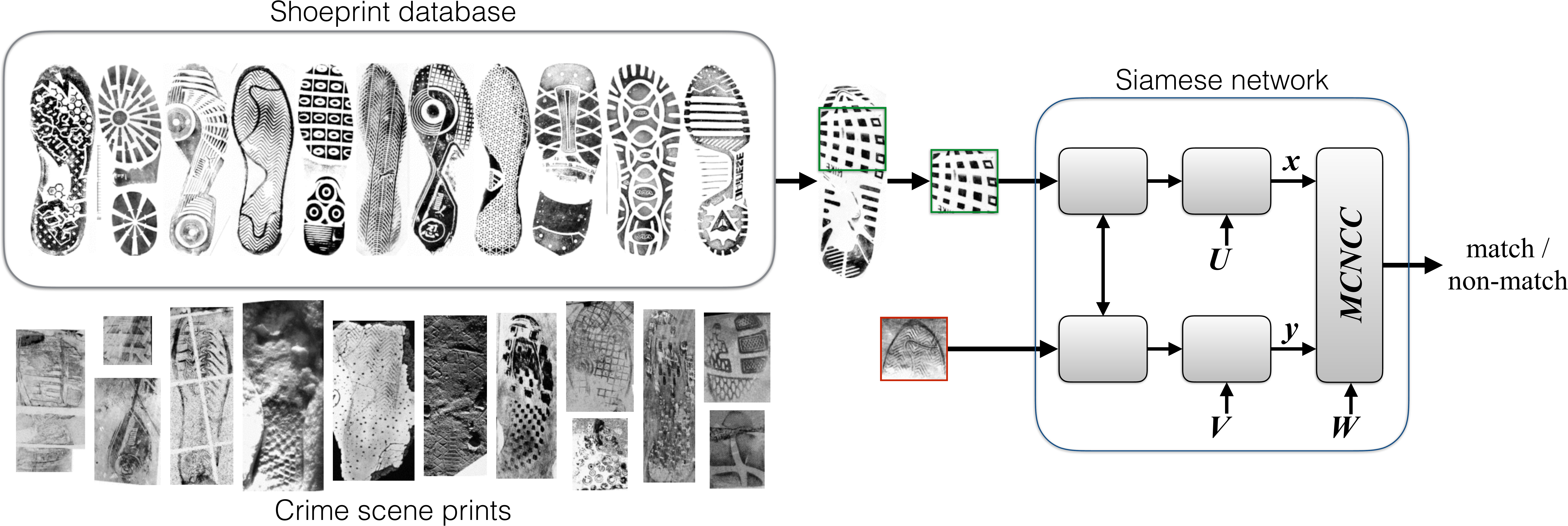}
  \end{tabular}\vspace{-2mm}
\end{center}
\caption{We would like to match crime scene prints to a database of test
impressions despite significant cross-domain differences in appearance. We
utilize a Siamese network to perform matching using a multi-channel normalized
cross correlation. We find that per-exemplar, per-channel normalization of
CNN feature maps significantly improves matching performance. Here $U$ and $V$
are the linear projection parameters for laboratory test impression and crime
scene photo domains respectively. $W$ is the per-channel importance weights.
And $x$ and $y$ are the projected features of each domain used for matching.}
  \label{fig:overview}
\end{figure*}

Analysis of shoe tread impressions is made difficult by the variability in
types of crime scene evidence (ranging from traces of dust or oil on hard
surfaces to impressions made in soil) and the lack of comprehensive datasets of
shoe outsole tread patterns  (see Fig.~\ref{fig:overview}). Solving this
problem requires developing models that can handle \textit{cross-domain}
matching of tread features between photos of clean test impressions (or images
of shoe outsoles) and photos of crime scene evidence.  We face the additional
challenge that we would like to use extracted image features for matching a
given crime scene impression to a large, open-ended database of exemplar tread
patterns.

Cross-domain image matching arises in a variety of other application domains
beyond our specific scenario of forensic shoeprint matching.  For example,
matching aerial photos to GIS map data for location
discovery~\cite{SenletICPR2014,CosteaBMVC2016,DivechaSIGSPATIAL16}, image
retrieval from hand drawn sketches and
paintings~\cite{ChenetalSketch2Photo2009,ShrivastavaCrossDomain2011},
and matching images to 3D models~\cite{RusselAlignment2011}.
As with shoeprint matching, many of these
applications often lack large datasets of ground-truth examples of cross-domain
matches. This lack of training data makes it difficult to learn cross-domain
matching metrics directly from raw pixel data.  Instead traditional approaches
have focused on designing feature extractors for each domain which yield domain
invariant descriptions (e.g., locations of edges) which can then be directly
compared.

Deep convolutional neural net (CNN) features hierarchies have proven incredibly
effective at a wide range of recognition tasks. Generic feature extractors
trained for general-purpose image categorization often perform surprising well
for novel categorization tasks without performing any fine-tuning beyond
training a linear classifier \cite{sharif2014cnn}.  This is
often explained by appealing to the notion that these learned representations
extract image features with invariances that are, in some sense, generic.  We
might hope that these same invariances would prove useful in our setting (\eg,
encoding the shape of a tread element in a way that is insensitive to shading,
contrast reversals, etc.). However, our problem differs in that we need to
formulate a cross-domain similarity metric rather than simply training a k-way
classifier. 

Building on our previous work~\cite{KongSRF_BMVC_2017},
we tackle this problem using similarity measures that are derived from normalized
cross-correlation (NCC), a classic approach for matching gray-scale templates.
For CNN feature maps, it is necessary to extend this to handle multiple channels. 
Our contribution is to propose a multi-channel variant of NCC which performs
normalization on a per-channel basis (rather than, e.g., per-feature volume).
We find this performs substantially better than related similarity measures
such as the widely used cosine distance. We explain this finding in terms of
the statistics of CNN feature maps. Finally, we use this multi-channel NCC as a
building block for a Siamese network model which can be trained end-to-end to
optimize matching performance.

\section{Related Work}


\paragraph{Shoeprint recognition} The widespread success of automatic
fingerprint identification systems (AFIS)~\cite{lee2001advances} has inspired
many attempts to similarly automate shoeprint recognition. Much initial work in
this area focused on developing feature sets that are rotation and
translation invariant.  Examples include, phase only
correlation~\cite{gueham2008automatic}, edge histogram DFT
magnitudes~\cite{zhang2005automatic}, power spectral
densities~\cite{de2005automated,dardi2009texture}, and the Fourier-Mellin
transform~\cite{gueham2008automatic}. Some other approaches pre-align the query
and database image using the Radon transform~\cite{patil2009rotation} while
still others sidestep global alignment entirely by computing only relative
features between keypoints pairs~\cite{tang2010footwear,pavlou2006automatic}.
Finally, alignment can be implicitly computed by matching rotationally
invariant keypoint descriptors between the query and database
images~\cite{pavlou2006automatic,wei2014alignment}.  The recent study of
Richetelli~\etal~\cite{Richetelli2017} carries out a comprehensive evaluation
of many of these approaches in a variety of scenarios using a carefully
constructed dataset of crime scene-like impressions.  In contrast to these
previous works, we handle global invariance by explicitly matching templates
using dense search over translations and rotations.

\paragraph{One-shot learning}
While we must match our crime scene evidence against a large database of
candidate shoes, our database contains very few examples per-class. As such, we
must learn to recognize each shoe category with as little as one training
example. This can be framed as a one-shot learning problem~\cite{li2006one}.
Prior work has explored one-shot object recognition with only a single training
example, or ``exemplar''~\cite{malisiewicz2011ensemble}.  Specifically in the
domain of shoeprints, Kortylewski~\etal~\cite{kortylewski2016probabilistic} fit
a compositional active basis model to an exemplar which could then be evaluated
against other images.  Alternatively, standardized or whitened off-the-shelf
HOG features have proven very effective for exemplar
recognition~\cite{hariharan2012discriminative}.  Our approach is similar in
that we examine the performance of one-shot recognition using generic deep
features which have proven surprisingly robust for a huge range of recognition
tasks~\cite{sharif2014cnn}.

\paragraph{Similarity metric learning}
While off-the-shelf deep features work well~\cite{sharif2014cnn}, they can be
often be {\em fine tuned} to improve performance on specific tasks. In
particular, for a paired comparison tasks, so-called ``Siamese'' architectures
integrate feature extraction and comparison in a single differentiable model
that can be optimized end-to-end.  Past work has demonstrated that Siamese
networks learn good features for person re-identification, face recognition,
and stereo matching~\cite{zbontar2015computing,parkhi2015deep,xiao2016learning}; 
deep pseudo-Siamese architectures can even learn to embed two dissimilar
domains into a common co-domain~\cite{zagoruyko2015learning}.  For shoe class
recognition, we similarly learn to embed two types of images: (1) crime scene
photos and (2) laboratory test impressions.

\section{Multi-variate Cross Correlation}
In order to compare two corresponding image patches, we extend the approach of
normalized cross-correlation (often used for matching gray-scale images) to work
with multi-channel CNN features.  Interestingly, there is not an immediately
obvious extension of NCC to multiple channels, as evidenced by multiple
approaches proposed in the literature
~\cite{fisher1995multi,martin1979multivariate,geiss1991multivariate,popper1974multivariate}.
To motivate our approach, we appeal to a statistical perspective. 

\paragraph{Normalized correlation} Let $x,y$ be two scalar random variables. A
standard measure of correlation between two variables is given by their {\em
Pearson's correlation coefficient}~\cite{martin1979multivariate}:
\begin{align}
  \rho(x,y) &= E[\tilde{x}\tilde{y}] = \frac{\sigma_{xy}}{\sqrt{\sigma_{xx}} \sqrt{\sigma_{yy}}} \label{eq:pearson} 
\end{align}
where
\begin{align}
  &\tilde{x} = \frac{x - \mu_x}{\sqrt{\sigma_{xx}}} \nonumber 
\end{align}
is the {\em standardized} version of $x$ (similarly for $y$) and 
\begin{align}
  &\mu_x = E[x] \nonumber \\
  &\sigma_{xx} = E[(x - \mu_x)^2] \nonumber \\
  &\sigma_{xy} = E[(x - \mu_x) (y - \mu_y)] \nonumber
\end{align}
Intuitively, the above corresponds to the correlation between two transformed
random variables that are ``whitened'' to have zero-mean and unit variance. The
normalization ensures that correlation coefficient will lie between $-1$ and
$+1$.

\paragraph{Normalized cross-correlation}  Let us model pixels $x$ from an image
patch $X$ as corrupted by some i.i.d. noise process and similarly pixels another
patch $Y$ (of identical size) as $y$.  The {\em sample} estimate of the
Pearson's coefficient for variables $x,y$ is equivalent to the normalized
cross-correlation (NCC) between patches $X,Y$:

{
\begin{align}
  \operatorname{NCC}(X,Y) = \frac{1}{|P|}\sum_{i \in P}
                            \frac{(x[i] - \mu_x)}{\sqrt{\sigma_{xx}}}
                            \frac{(y[i] - \mu_y)}{\sqrt{\sigma_{yy}}}
\end{align}}
where $P$ refers to the set of pixel positions in a patch and means and
standard deviations are replaced by their sample estimates.

From the perspective of detection theory, normalization is motivated by the
need to compare correlation coefficients across different pairs of samples with
non-stationary statistics (\eg, determining which patches $\{Y^1,Y^2,\ldots\}$
are the same as a given template patch $X$ where statistics vary from one $Y$
to the next). Estimating first and second-order statistics per-patch provides a
convenient way to handle sources of ``noise'' that are approximately i.i.d.
conditioned on the choice of patch $P$ but not independent of patch location.

\paragraph{Multivariate extension} Let us extend the above formulation for random
{\em vectors} ${\bf x}, {\bf y} \in R^N$ where $N$ corresponds to the
multiple channels of values at each pixel (\eg, $N=3$ for a RGB image). The
scalar correlation is now replaced by a $N \times N$ correlation {\em matrix}.
To produce a final score capturing the overall correlation, we propose to
use the {\em trace} of this matrix, which is equivalent to the sum of its
eigenvalues. As before, we add invariance by computing correlations on
transformed variables ${\bf \tilde{x},\tilde{y}}$ that are ``whitened'' to have
a zero-mean and identity covariance matrix:

\begin{align}
  \rho_{multi}({\bf x},{\bf y}) &= \frac{1}{N} Tr( E[{\bf \tilde{x} \tilde{y}}^T]) \label{eq:multi}\\
    &= \frac{1}{N}Tr( \Sigma_{\bf xx}^{-\frac{1}{2}} \Sigma_{\bf xy} \Sigma_{\bf yy}^{-\frac{1}{2}}) \nonumber
\end{align}
where:
\begin{align}
  \nonumber{\bf \tilde{x}} &= \Sigma_{\bf xx}^{-\frac{1}{2}}({\bf x} - \mu_{\bf x}), \\
  \nonumber\Sigma_{\bf xx} &= E[({\bf x} - {\bf \mu_x}) ({\bf x} - {\bf \mu_x})^T], \\
  \nonumber\Sigma_{\bf xy} &= E[({\bf x} - {\bf \mu_x}) ({\bf y} - {\bf \mu_y})^T].
\end{align}
The above multivariate generalization of the Pearson's coefficient is arguably
rather natural, and indeed, is similar to previous formulations that also make use
of a trace operator on a correlation
matrix~\cite{martin1979multivariate,popper1974multivariate}. However, one
crucial distinction from such past work is that our generalization
\eqref{eq:multi} reduces to \eqref{eq:pearson} for $N=1$. In particular,
~\cite{martin1979multivariate,popper1974multivariate} propose multivariate
extensions that are restricted to return a nonnegative coefficient. It is
straightforward to show that our multivariate coefficient will lie between $-1$
and $+1$.

\begin{figure}[t]
\begin{center}
\begin{tabular}{c}
  \includegraphics[trim=0 0 0 0,clip, width=80mm]{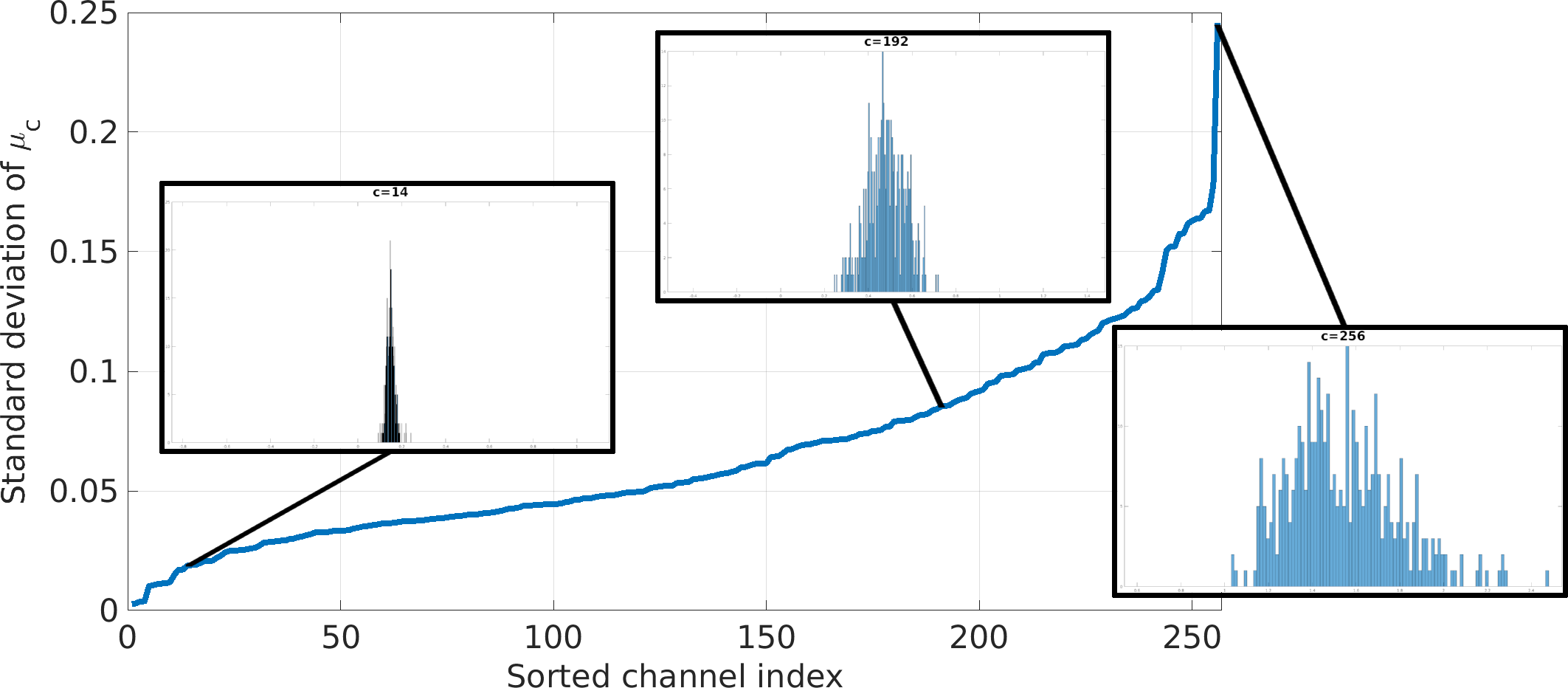}
  \end{tabular}\vspace{-2mm}
  \end{center}
  \caption{\textbf{Distribution of patch channel means: } 
    For each query image (patch) we match against the database, our proposed
    MCNCC similarity measure normalizes ResNet-50 `res2x' feature channels by
    their individual mean and standard deviation.  For uniformly sampled
    patches, we denote the normalizing mean for channel $c$ using the random
    variable $\mu_c$. For each channel, we plot the standard deviation of
    $\mu_c$ above with channels sorted by increasing standard deviation. When
    the mean response for a channel varies little from one patch to the next
    (small std, left), we can expect that a global, per-dataset transformation
    (\eg, PCA or CCA whitening) is sufficient to normalize the channel response.
    However, for channels where individual patches in the dataset have very
    different channel means (large std, right), normalizing by the local
    (per-patch) statistics provides additional invariance.}
  \label{fig:std_of_mu_histograms}
\end{figure}

\paragraph{Decorrelated channel statistics} The above formulation can be
computationally cumbersome for large $N$, since it requires obtaining sample
estimates of matrices of size $N^2$. Suppose we make the strong assumption that
all $N$ channels are {\em uncorrelated} with each other. This greatly
simplifies the above expression, since the covariance matrices are then
diagonal matrices:
\begin{align*}
\Sigma_{\bf xy} = \operatorname{diag}(\{\sigma_{x_cy_c}\}) \\
\Sigma_{\bf xx} = \operatorname{diag}(\{\sigma_{x_cx_c}\}) \\
\Sigma_{\bf yy} = \operatorname{diag}(\{\sigma_{y_cy_c}\})
\end{align*}
Plugging this assumption into \eqref{eq:multi} yields the simplified expression
for multivariate correlation
\begin{align}
 \rho_{multi}({\bf x},{\bf y}) = \frac{1}{N} \sum_{c=1}^N \frac{\sigma_{x_cy_c}}{\sqrt{\sigma_{x_cx_c}} \sqrt{\sigma_{y_cy_c}}} \label{eq:diag}
\end{align}
where the diagonal multivariate statistic is simply the average of $N$ per-channel
correlation coefficients. It is easy to see that this sum must lie between $-1$
and $+1$.

\paragraph{Multi-channel NCC}  The sample estimate of \eqref{eq:diag} yields a
multi-channel extension of NCC which is adapted to the patch:
{\small \begin{align}
  \nonumber \operatorname{MCNCC}(X,Y) = \frac{1}{N|P|} \sum_{c=1}^N\sum_{i \in P}
                              \frac{(x_c[i] - \mu_{x_c})}{\sqrt{\sigma_{x_cx_c}}}
                              \frac{(y_c[i] - \mu_{y_c})}{\sqrt{\sigma_{y_cy_c}}}
\end{align}}
The above multi-channel extension is similar to the final formulation
in~\cite{fisher1995multi}, but is derived from a statistical assumption
on the channel correlation.

\begin{figure}[t]
\begin{center}
\begin{tabular}{c}
  \includegraphics[trim=0 0 0 0,clip, width=80mm]{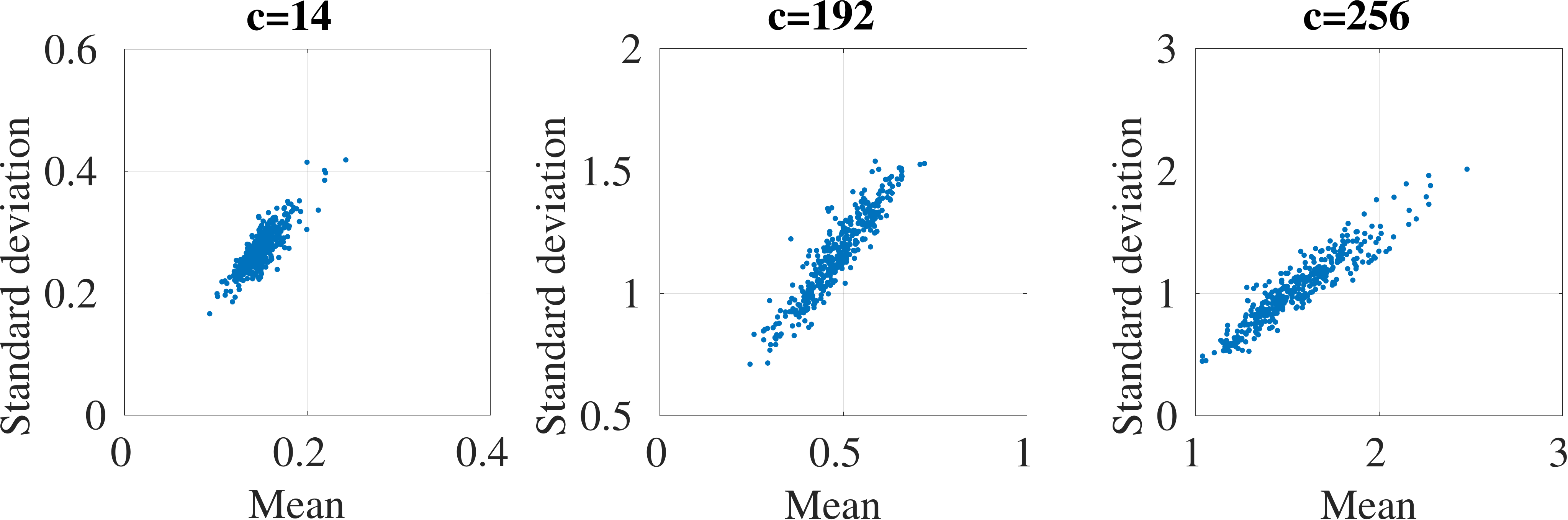}
  \end{tabular}\vspace{-2mm}
  \end{center}
  \caption{\textbf{Normalizing channel statistics: } As shown in the histograms
  of Fig.~\ref{fig:std_of_mu_histograms}, for some feature channels, patches
  have wildly different means and standard deviations. For channel 14 (left),
  the statistics (and hence normalization) are similar from one patch to the
  next while for channel 256 (right), means and standard deviations vary
  substantially across patches. CNN channel activations are positive so means
  and standard deviations are strongly correlated.}
  \label{fig:mean_vs_std}
\end{figure}

\paragraph{Cross-domain covariates and whitening} Assuming a diagonal covariance
makes strong assumptions about cross-channel correlations.  When strong
cross-correlations exist, an alternative approach to reducing computational
complexity is to assume that cross-channel correlations lie within a $K$
dimensional subspace, where $K \leq N$. We can learn a projection matrix for
reducing the dimensionality of features from both patch $X$ and $Y$ which
decorrelates and scales the channels to have unit variance:

{\small
\begin{align*}
{\bf \hat{x}} = U ({\bf x} - \mu_x), \quad U \in R^{K \times N}, \quad E[{\bf \hat{x}} {\bf \hat{x}}^T] = I\\
{\bf \hat{y}} = V ({\bf y} - \mu_y), \quad V \in R^{K \times N}, \quad E[{\bf \hat{y}} {\bf \hat{y}}^T] = I
\end{align*}}
In general, the projection matrix could be different for different domains (in
our case, crime scene versus test prints). One strategy for learning the projection
matrices is applying principle component analysis (PCA) on samples from each domain
separately. Alternatively, when paired training examples are available, one
could use canonical correlation analysis (CCA)~\cite{MardiaKentBibby1980}, which
jointly learn the projections that maximize correlation across domains.
An added benefit of using {\em orthogonalizing}
transformations such as PCA/CCA is that transformed data satisfies the diagonal
assumptions (globally) allowing us to estimate patch multivariate correlations
in this projected space with diagonalized covariance matrices of size $K \times
K$. 

\paragraph{Global versus local whitening}
There are two distinct aspects to whitening (or normalizing) variables in our
problem setup to be determined: (1) assumptions on the structure of the sample
mean and covariance matrix, and (2) the data over which the sample mean and
covariance are estimated. In choosing the structure, one could enforce an
unrestricted covariance matrix, a low-rank covariance matrix (e.g., PCA), or a
diagonal covariance matrix (e.g., estimating scalar means and variances).  In
choosing the data, one could estimate these parameters over individual patches
(local whitening) or over the entire dataset (global whitening). In
Section~\ref{sec:diag_experiments}, we empirically explore various combinations
of these design choices which are computationally feasible (e.g., estimating a
full-rank covariance matrix locally for each patch would be too expensive).  We
find a good tradeoff to be global whitening (to decorrelate features globally),
followed by local whitening with a diagonal covariance assumption (e.g.,
MCNCC).


To understand the value of global and per-patch normalization, we examine the
statistics of CNN feature channels across samples of our dataset.
Fig.~\ref{fig:std_of_mu_histograms} and Fig.~\ref{fig:mean_vs_std} illustrate
how the per-channel normalizing statistics ($\mu_c,\sigma_c$) vary across
patches and across channels. Notably, for some channels, the normalizing
statistics change substantially from patch to patch. This makes the results of
performing local, per-patch normalization significantly different from global,
per-dataset normalization.  

One common effect of both global and local whitening is to prevent feature
channels that tend to have large means and variances from dominating the
correlation score. However, by the same merit this can have the undesirable
effect of amplifying the influence of low-variance channels which may not be
discriminative for matching.  In the next section we generalize both PCA and
CCA using a learning framework which can learn channel decorrelation and
per-channel importance weighting by optimizing a discriminative performance
objective.

\section{Learning Correlation Similarity Measures}
\label{sec:learning}

In order to allow for additional flexibility of weighting the relevance of each
channel we consider a channel-weighted variant of MCNCC parameterized by vector
$W$:

\vspace{-0.05in}
{\scriptsize
\begin{align}
  \nonumber
  \operatorname{MCNCC}&_{W}(X,Y) \\
    &= \sum_{c=1}^N  W_c \left[
    \frac{1}{|P|} \sum_{i \in P}
    \frac{(x_c[i] - \mu_{x_c})}{\sqrt{\sigma_{x_cx_c}}}
    \frac{(y_c[i] - \mu_{y_c})}{\sqrt{\sigma_{y_cy_c}}}
    \right]
\end{align}}
This per-channel weighting can undo the effect of scaling by the standard
deviation in order to re-weight channels by their informativeness.
Furthermore, since the features $x,y$ are themselves produced by a CNN model,
we can consider the parameters of that model as additional candidates for
optimization.  In this view, PCA/CCA can be seen as adding an extra linear
network layer prior to the correlation calculation.  The parameters of such a
layer can be initialized using PCA/CCA and then discriminatively tuned. The
resulting ``Siamese'' architecture is illustrated in Fig.~\ref{fig:overview}.

\paragraph{Siamese loss:} To train the model, we minimize a hinge-loss:
\begin{align}
  &\underset{W,U,V,b}{\arg\min} \quad\frac{\alpha}{2}\|W\|_2^2
    + \frac{\beta}{2}\left( \|U\|_F^2 + \|V\|_F^2 \right) \\\nonumber
  &+ \sum_{s,t} \operatorname{max}\left( 0,
    1 - z_{s,t}\operatorname{MCNCC}_{W}(\phi_{U}(X^s),\phi_{V}(Y^t)) + b
               \right)
\end{align}
where we have made explicit the function $\phi$ which computes the deep
features of two shoeprints $X^s$ and $Y^t$, with $W$, $U$, and $V$ representing
the parameters for the per-channel importance weighting and the linear
projections for the two domains respectively.
$b$ is the bias and $z_{s,t}$ is a binary same-source label (\ie, $+1$ when $X^s$ and
$Y^t$ come from the same source and $-1$ otherwise).
Finally, $\alpha$ is the regularization hyperparameter for $W$ and $\beta$ is
the same for $U$ and $V$.

We implement $\phi$ using a deep architecture, which is trainable using
standard backpropagation. Each channel contributes a term to the MCNCC
which itself is just a single channel (NCC) term.  The operation is symmetric in
$X$ and $Y$, and the gradient can be computed efficiently by reusing the NCC
computation from the forward pass:

{\scriptsize
\begin{equation}
  \frac{\total\operatorname{NCC}(x_c,y_c)}{\total x_c[j]} =
  \frac{1}{|P|\sqrt{\sigma_{x_cx_c}}}
  ( \tilde{y}_c[j]+ \tilde{x}_c[j] \operatorname{NCC}(x_c,y_c) )
\end{equation}}

\begin{figure*}[t]
  \begin{center}
  \begin{tabular}{cc}
    \includegraphics[trim=0 0 0 0,clip, height=50mm]{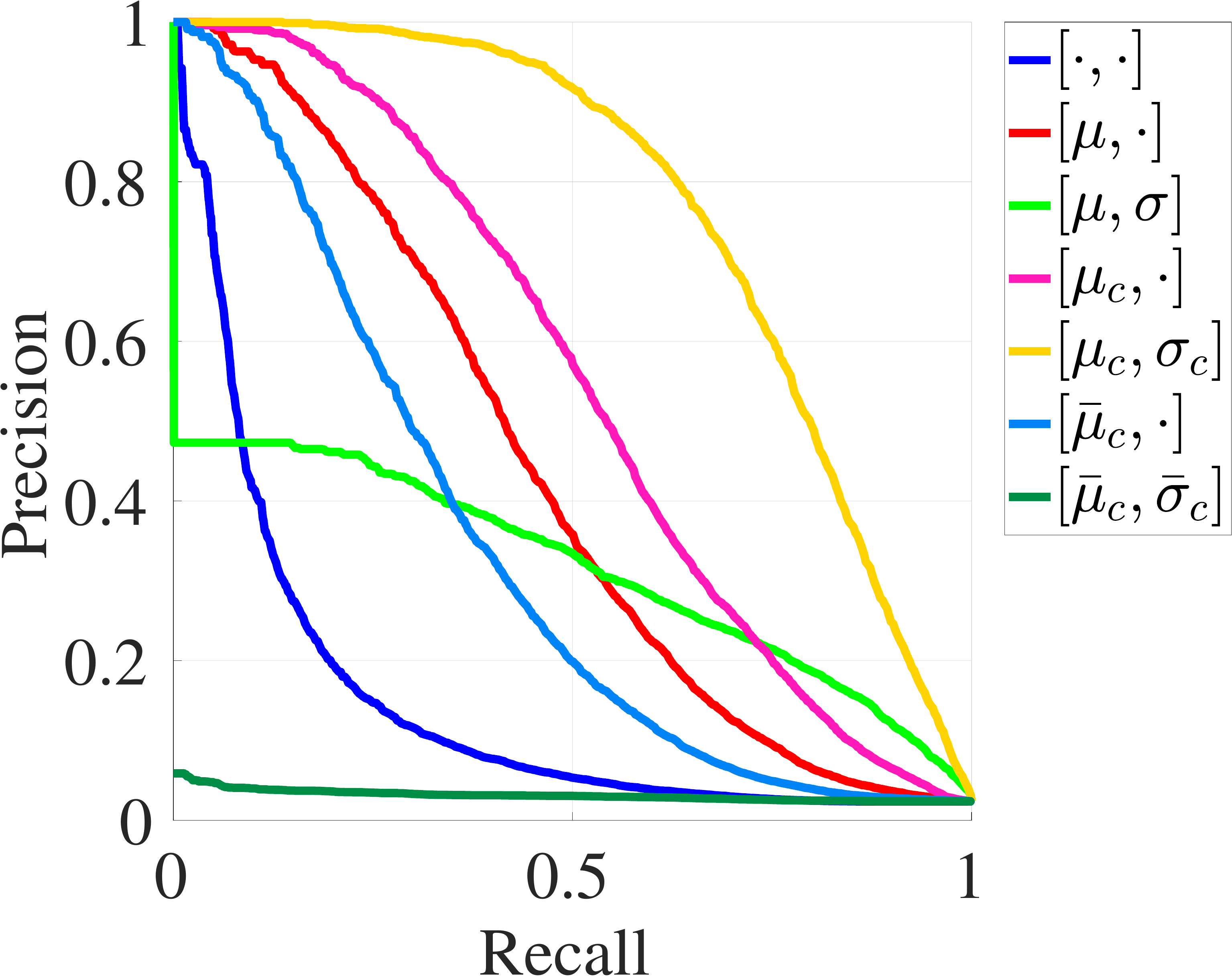} &
    \includegraphics[trim=0 0 0 0,clip, height=50mm]{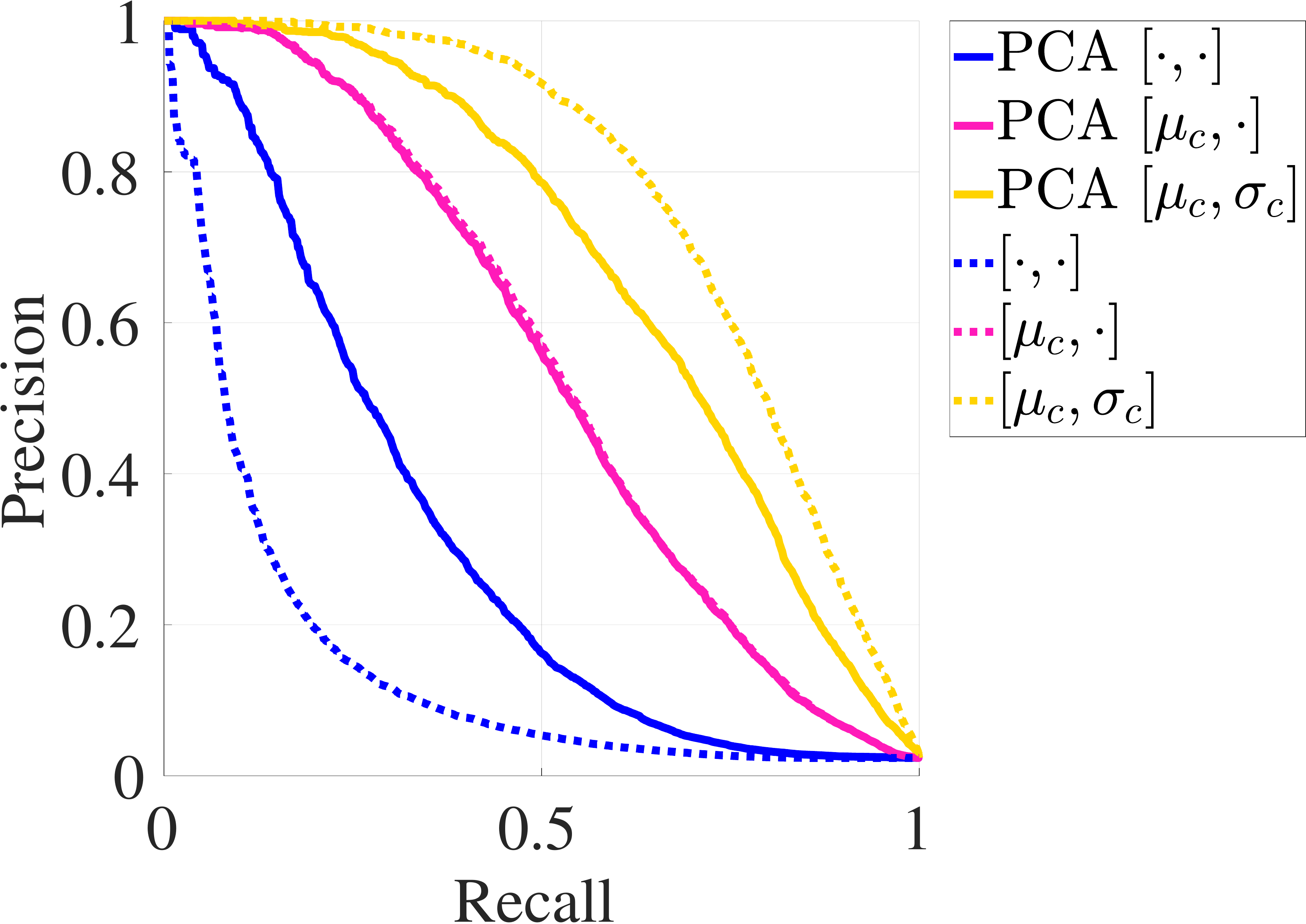}
  \end{tabular}\vspace{-2mm}
  \end{center}
  \caption{Comparing MCNCC to baselines for image retrieval within the same
  domain. The methods are denoted by two
  operations in square brackets: centering and normalization,
  respectively. $\mu$ and $\sigma$ denote computing the statistics across all
  channels, $\mu_c$ and $\sigma_c$ denote computing per-channel statistics, and
  $\cdot$ denotes the absence of the operation (\eg, MCNCC is denoted as
  $[\mu_c,\sigma_c]$, whereas cross-correlation is denoted as $[\cdot,\cdot]$.
  Finally, $\bar{\mu}_c$ and $\bar{\sigma}_c$ denote computing the average
  per-channel statistics across the dataset.
  The left panel shows the performance on the raw features, whereas the
  right panel compares globally whitened features using PCA (solid lines) against
  their corresponding raw features (dotted lines). (Best viewed in color.)}
  \label{fig:ncc_ablation}
\end{figure*}

\paragraph{Derivation of NCC gradient:} To derive the NCC gradient, we first
expand it as a sum over individual pixels indexed by $i$ and consider the total
derivative with respect to input feature $x[j]$

\vspace{-0.05in}
{\scriptsize
\begin{align}
\nonumber & \frac{\total \operatorname{NCC}(x,y)}{\total x[j]} = \\
  & \qquad \qquad \frac{1}{|P|}\sum_{i \in P}
  \tilde{y}[i] \left(\frac{\partial \tilde{x}[i]}{\partial x[j]}+
  \frac{\partial \tilde{x}[i]}{\partial \mu_x}\frac{\partial \mu_x}{\partial x[j]} +
  \frac{\partial \tilde{x}[i]}{\partial \sigma_{xx}}\frac{\partial \sigma_{xx}}{\partial x[j]} \right)
  \label{eqn:totalderiv}
\end{align}}
where we have have dropped the channel subscript for clarity. 
The partial derivative $\tfrac{\partial \tilde{x}[i]}{\partial x[j]} =
\tfrac{1}{\sqrt{\sigma_{xx}}}$, if and only if $i=j$ and is zero otherwise. 
The remaining partials derive as follows:
{\scriptsize
\begin{align*}
  \frac{\partial \tilde{x}[i]}{\partial \mu_x} &= -\frac{1}{\sqrt{\sigma_{xx}}} \quad 
  & \frac{\partial \mu_x}{\partial x[j]} &= \frac{1}{|P|} \\ 
  \frac{\partial \tilde{x}[i]}{\partial \sigma_{xx}} &= \frac{1}{2 \sigma_{xx}^{3/2}} (x[i]-\mu_x) \quad
  & \frac{\partial \sigma_{xx}}{\partial x[j]}  &= \frac{2 \left(x[j] - \mu_x \right)}{|P|}
\end{align*}}
Substituting them into Eq.~\ref{eqn:totalderiv}, we arrive at a final
expression:
{\scriptsize
\begin{align}
  &\frac{\total\operatorname{NCC}(x,y)}{\total x[j]} \nonumber \\ \nonumber 
  &= \frac{\tilde{y}[j]}{|P|\sqrt{\sigma_{xx}}} +
  \frac{1}{|P|}\sum_{i \in P} \tilde{y}[i] \left(
  \frac{-1}{|P|\sqrt{\sigma_{xx}}} +
  \frac{2\left(x[i]-\mu_x \right) \left(x[j]-\mu_x \right)}{2 |P| \sigma_{xx}^{3/2}} \right) \\\nonumber
  &= \frac{1}{|P| \sqrt{\sigma_{xx}}} \left (
  \tilde{y}[j] + 
  \frac{1}{|P|} \sum_{i \in P} \tilde{y}[i] \left( -1 + \frac{\left(x[i]-\mu_x \right) \left(x[j]-\mu_x \right)}{ \sigma_{xx}} \right)
  \right ) \\\nonumber
  &= \frac{1}{|P| \sqrt{\sigma_{xx}}} \left (
  \tilde{y}[j] - 
  \frac{1}{|P|} \sum_{i \in P} \tilde{y}[i] +
 \frac{1}{|P|} \sum_{i \in P} \tilde{y}[i] \tilde{x}[i] \tilde{x}[j] \right) \\
  &= \frac{1}{|P|\sqrt{\sigma_{xx}}} ( \tilde{y}[j]+ \tilde{x}[j] \operatorname{NCC}(x,y) )
\end{align}}
where we have made use of the fact that $\tilde{y}$ is zero-mean.

\section{Diagnostic Experiments}
\label{sec:diag_experiments}
To understand the effects of feature channel normalization on retrieval
performance, we compare the proposed MCNCC measure to two baseline approaches:
simple unnormalized cross-correlation and cross-correlation normalized by a
single $\mu$ and $\sigma$ estimated over the whole 3D feature volume. We note
that the latter is closely related to the ``cosine similarity'' which is
popular in many retrieval applications (cosine similarity scales by $\sigma$
but does not subtract $\mu$). We also consider variants which only perform
partial standardization and/or whitening of the input features.

\paragraph{Partial print matching:}
We evaluate these methods in a setup that mimics the occurrence of partial
occlusions in shoeprint matching, but focus on a single modality of test
impressions.  We extract 512 query patches (random selected $97\times97$ pixel
sub-windows) from test impressions that have two or more matching tread
patterns in the database.  The task is then to retrieve from the database the
set of relevant prints.  As the query patches are smaller than the test
impressions, we search over spatial translations (with a stride of 1), using
the maximizing correlation value to score the match to the test impression. We
do not need to search over rotations as all test impressions were aligned to a
canonical orientation.  When querying the database, the original shoeprint the
query was extracted from is removed (\ie, the results do not include the
self-match).

We carry out these experiments using a dataset that contains 387 test
impression of shoes and 137 crime scene prints collected by the Israel National
Police~\cite{yekutieli2012expert}.  As this dataset is not publicly available,
we used this dataset primarily for the diagnostic analysis and for training and
validating learned models.  In these diagnostic experiments, except where noted 
otherwise, we use the
256-channel `res2bx' activations from a pre-trained ResNet-50
model\footnote{Pretrained model was obtained from
\url{http://www.vlfeat.org/matconvnet/models/imagenet-resnet-50-dag.mat}}.  We
evaluated feature maps at other locations along the network, but found those to
performed the best.
\begin{figure*}[t]
\begin{center}
\begin{tabular}{cc}
  \includegraphics[trim=0 0 0 0,clip, width=80mm]{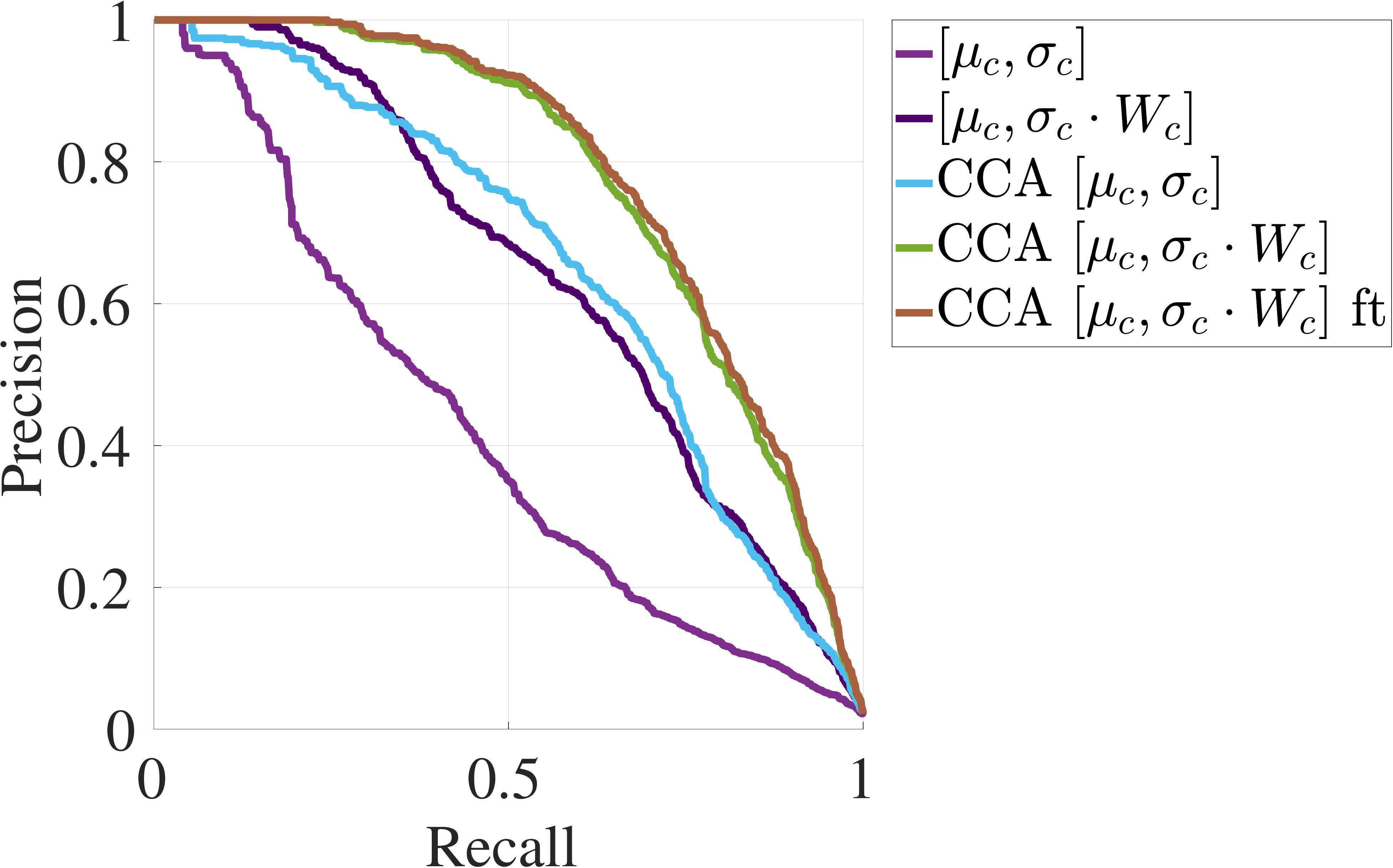}
  \includegraphics[trim=0 0 0 0,clip, width=80mm]{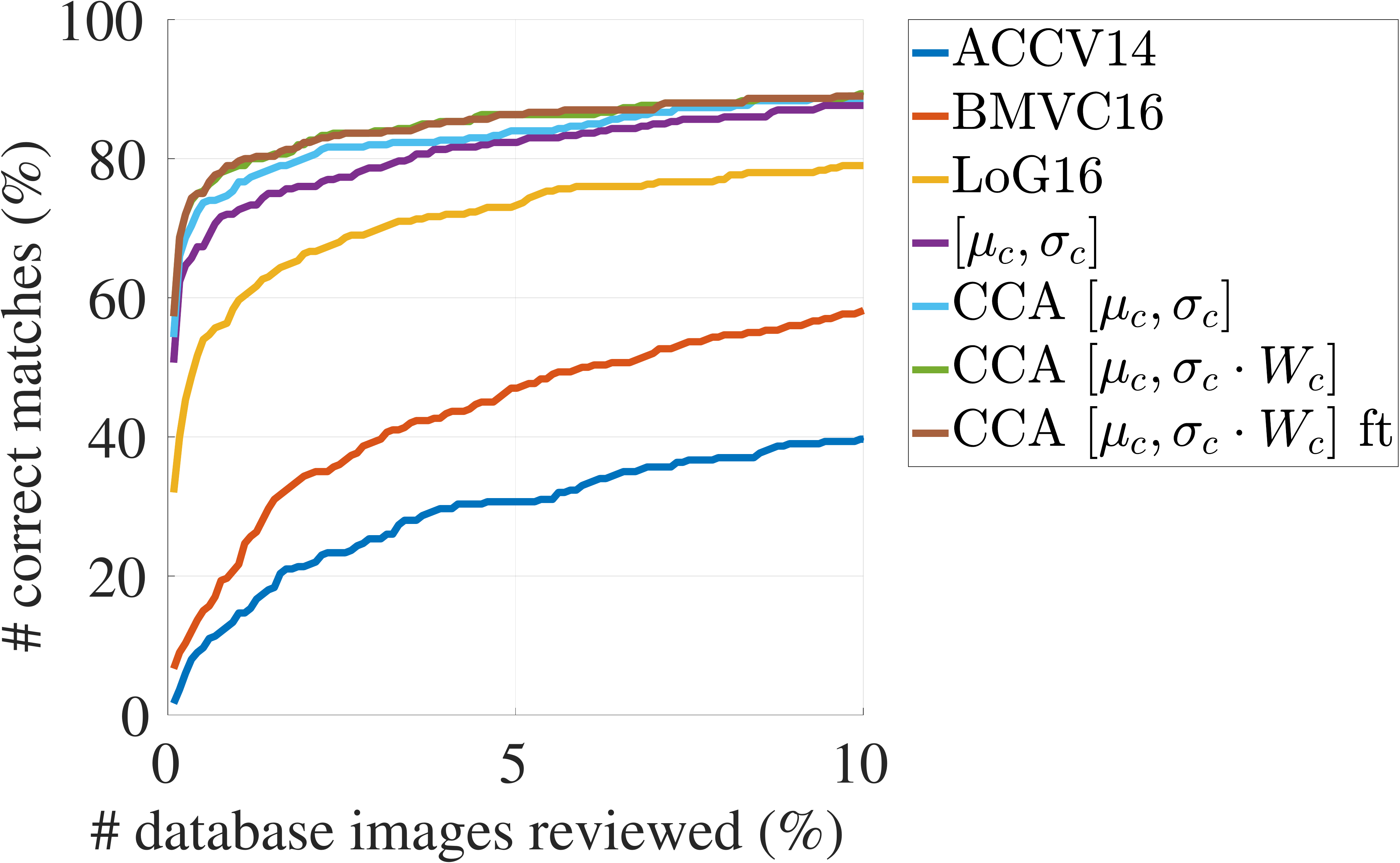}
  \end{tabular}\vspace{-2mm}
  \end{center}
  \caption{Comparing MCNCC with uniform weights (denoted as $[\mu_c,\sigma_c]$),
  learned per-channel weights (denoted as $[\mu_c,\sigma_c\cdot W_c]$),
  learned linear projections (denoted as CCA~$[\mu_c,\sigma_c]$), piece-wise
  learned projection and per-channel weights (denoted as
  CCA~$[\mu_c,\sigma_c\cdot W_c]$), and jointly learned projection and
  per-channel
  weights (denoted as CCA~$[\mu_c,\sigma_c\cdot W_c]$~ft) for
  retrieving relevant shoeprint test impressions for crime scene prints.
  The left panel shows our five methods on the Israeli dataset.
  The right panel compares variants of our proposed system against
  the current state-of-the-art, as published in:
  ACCV14~\cite{kortylewski2014unsupervised},
  BMVC16~\cite{kortylewski2016probabilistic} and LoG16~\cite{kortylewski2017model}
  using cumulative match characteristic (CMC).}
  \label{fig:latent2trace}
\end{figure*}

\paragraph{Global versus local normalization:}
Fig.~\ref{fig:ncc_ablation} shows retrieval performance in terms of the
tradeoff of precision and recall at different match thresholds.
In the legend we denote different schemes in
square brackets, where the first term indicates the centering operation and the
second term indicates the normalization operation.  A $\cdot$ indicates the
absence of the operation.  $\mu$ and $\sigma$ indicate that standardization 
was performed using local (\ie, \emph{per-exemplar}) statistics of features over
the entire (3D) feature map.  $\mu_c$ and $\sigma_c$ indicate local per-channel
centering and normalization.  $\bar{\mu}_c$ and $\bar{\sigma}_c$ indicate
global per-channel centering and normalization (\ie, statistics are estimated
over the \emph{whole dataset}).  Therefore, simple unnormalized
cross-correlation is indicated as $[\cdot,\cdot]$, cosine distance is indicated
as $[\mu,\sigma]$, and our proposed MCNCC measure is indicated as
$[\mu_c,\sigma_c]$.

We can clearly see from the left panel of Fig.~\ref{fig:ncc_ablation} that using
per-channel statistics estimated independently for each comparison gives
substantial gains over the baseline methods.  Centering using 3D
(across-channel) statistics is better than either centering using global
statistics or just straight correlation.  But cosine distance (which adds the
scaling operation) decreases performance substantially for the low recall
region. In general, removing the mean response is far more important than
scaling by the standard deviation.  Interestingly, in the case of cosine
distance and global channel normalization, scaling by the standard deviation
actually hurts performance (\ie, $[\mu,\sigma]$ versus $[\mu,\cdot]$ and
$[\bar{\mu}_c,\bar{\sigma}_c]$ versus $[\bar{\mu}_c,\cdot]$ respectively). As
normalization re-weights channels, we posit that this may be negatively
effecting the scores by down-weighing important signals or boosting noisy
signals.

\paragraph{Channel decorrelation:} Recall that, for efficiency reasons, our
multivariate estimate of correlation assumes that channels are largely
decorrelated.  We also explored decorrelating the channels globally using a
full-dimension PCA (which also subtracts out the global mean $\bar{\mu}_c$).
The right panel of Fig.~\ref{fig:ncc_ablation} shows a comparison of these
decorrelated feature channels (solid curves) relative to baseline ResNet
channels (dotted curves). While the decorrelated features outperform baseline
correlation (due to the mean subtraction) we found that full MCNCC on the raw
features performed better than on globally decorrelated features. This may be
explained in part due to the fact that decorrelated features show an even wider
range of variation across different channels which may exacerbate some of the
negative effects of scaling by $\sigma_c$.

\paragraph{Other feature extractors:}
To see if this behavior was specific to the ResNet-50 model, we evaluate on
three additional features: raw pixels, GoogleNet, and DeepVGG-16. From the
GoogleNet model\footnote{Pretrained model was obtained from
\url{http://www.vlfeat.org/matconvnet/models/imagenet-googlenet-dag.mat}} we
used the 192-channel `conv2x' activations, and from the DeepVGG-16
model\footnote{Pretrained model was obtained from
\url{http://www.vlfeat.org/matconvnet/models/imagenet-vgg-verydeep-16.mat}} we
used the 256-channel `x12' activations.
We chose these particular CNN feature maps because they had the same or similar
spatial resolution as `res2bx' and were the immediate output of a rectified
linear unit layer.

As shown in Table~\ref{tab:feature_ablation}, we see a similar pattern to what we
observed with ResNet-50's `res2bx' features. Namely, that straight
cross-correlation (denoted as $[\cdot,\cdot]$) performs poorly, while MCNCC
(denoted as $[\mu_c,\sigma_c]$) performs the best. One significant departure
from the previous results for `res2bx' features is how models using entire
feature volume statistics perform. Centering using 3D statistics (denoted as
$[\mu,\cdot]$) yields performance that is closer to straight correlation, on the
other hand, standardizing using 3D statistics (denoted as $[\mu,\sigma]$) yields
performance that is closer to MCNCC when using GoogleNet's `conv2x' and
DeepVGG-16's `x12' features. 

When we look at the difference between the per-channel and the across-channel
(3D) statistics for query patches, we observe significant difference in sparsity
of $\mu_c$ compared to $\mu$: `conv2x' is about 2x more sparse than `x12,' which
itself is about 2x more sparse than `res2bx.' The level of sparsity correlates
with the performance of $[\mu,\cdot]$ compared to straight correlation across
the different features. The features where $\mu_c$ is more sparse, using $\mu$
overshifts across more channels leading to less performance gain relative to
straight correlation. When we look at the difference between
$\sigma$ and $\sigma_c$, we observe that $\sigma$ is on average larger than
$\sigma_c$. This means that compared to $\sigma_c$, using $\sigma$ dampens the
effect of noisy channels rather than boosting them. Looking at the change of
performance from $[\mu,\cdot]$ to $[\mu,\sigma]$ for different features, we
similarly see improvement roughly correlates to how much larger $\sigma$ is
than $\sigma_c$.

\begin{table*}[t]
\begin{center}
\begin{tabular}{c|c|c|c|c|c}
  Features & $[\cdot,\cdot]$ & $[\mu,\cdot]$ & $[\mu,\sigma]$ & $[\mu_c,\cdot]$
  & $[\mu_c,\sigma_c]$ \\\hline
  Raw Pixels & 0.04 & 0.20 & 0.45 & - & - \\
  ResNet-50 (res2bx) & 0.15 & 0.44 & 0.32 & 0.55 & 0.77 \\
  GoogleNet (conv2x) & 0.07 & 0.09 & 0.68 & 0.61 & 0.81 \\
  DeepVGG-16 (x20) & 0.09 & 0.31 & 0.73 & 0.51 & 0.76
\end{tabular}
\end{center}
\caption{Ablation study on the two normalized cross-correlation schemes across
  different features.  We measure performance using mean average precision,
  higher is better.  As the images are gray-scale single-channel images, for
  raw pixels $[\mu,\cdot]$ and $[\mu,\sigma]$ are identical to $[\mu_c,\cdot]$
  and $[\mu_c,\sigma_c]$, respectively.}
\label{tab:feature_ablation}
\end{table*}

\section{Cross-Domain Matching Experiments}
In this section, we evaluate our proposed system in settings that closely
resembles various real-world scenarios where query images are matched to a
database containing images from a different domain than that of the query.
We focus primarily on matching crime scene prints to a collection of test
impressions, but also demonstrate the effectiveness of MCNCC on two other
cross-domain applications: semantic segmentation label retrieval from building
facade images, and map retrieval from aerial photos.\footnote{Our code is
available at \url{http://github.com/bkong/MCNCC}} As in our diagnostic
experiments, we use the same pre-trained ResNet-50 model.  We use the
256-channel `res2bx' activations for the shoeprint and building facade data,
but found that the 1024-channel `res4cx' activations performed better
for the map retrieval task.

\begin{figure*}[t]
\begin{center}
  \includegraphics[trim=0 0 0 0,clip, width=160mm]{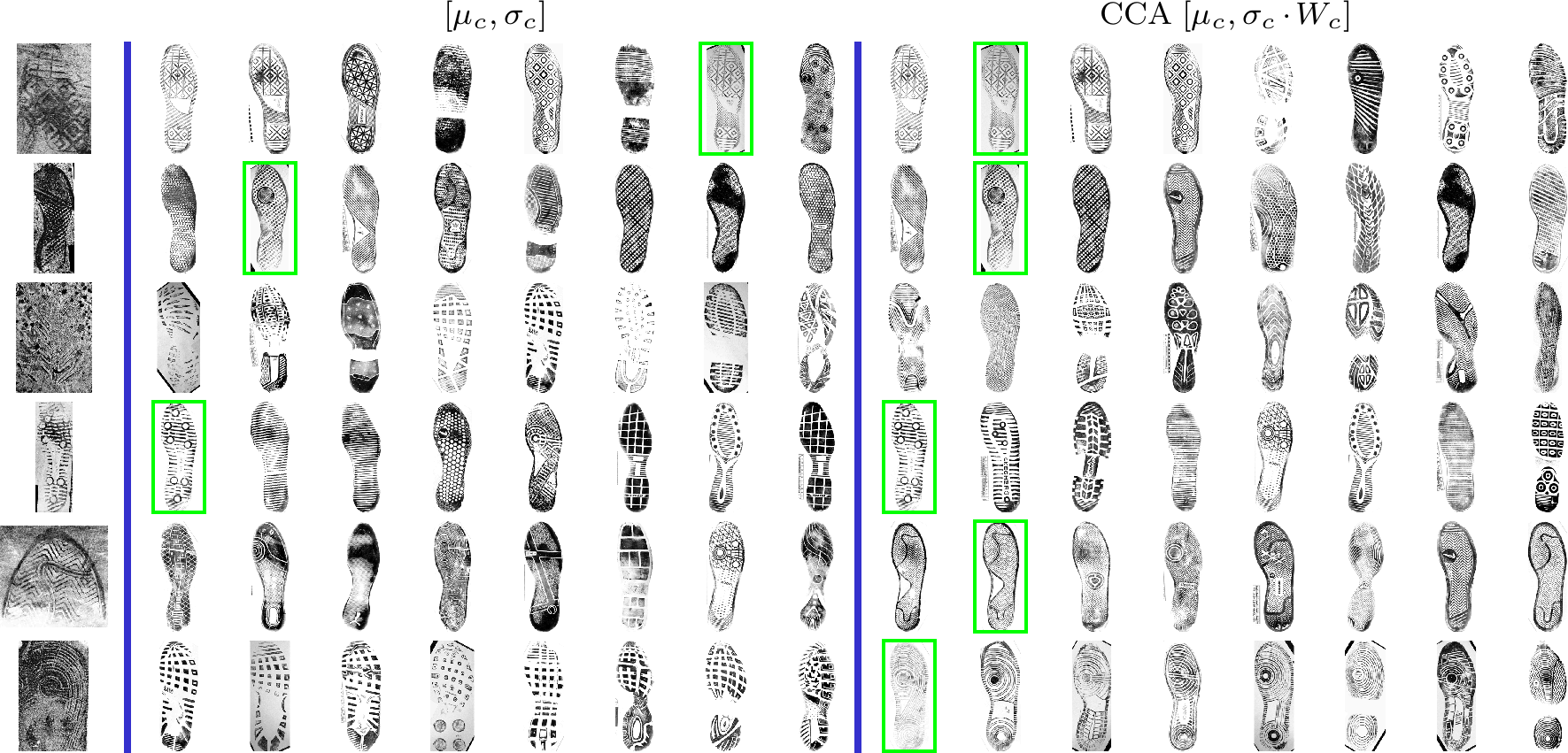}
  \vspace{-2mm}
\end{center}
\caption{FID-300 retrieval results. The left column shows the query crime scene
  prints, the middle column shows the top-8 results for $[\mu_c,\sigma_c]$, and
  the right column shows the top-8 results for CCA~$[\mu_c,\sigma_c\cdot W_c]$.
  Green boxes indicate the corresponding ground truth test impression.}
\label{fig:basel_retrieval}
\end{figure*}

\subsection{Shoeprint Retrieval}
In addition to the internal dataset described in
Section~\ref{sec:diag_experiments}, we also evaluated our approach on a
publicly available benchmark, the footwear identification dataset
(FID-300)~\cite{kortylewski2014unsupervised}.  FID-300 contains 1175 test
impressions and 300 crime scene prints.  The task here is similar to the
diagnostic experiments on patches, but now matching whole prints across
domains.  As the crime scene prints are not aligned to a canonical orientation,
we search over both translations (with a stride of 2) and rotations (from
-20$^\circ$ to +20$^\circ$ with a stride of 4$^\circ$). For a given alignment,
we compute the valid support region $P$ where the two images overlap. The local
statistics and correlation is only computed within this region.

As mentioned in Sec.~\ref{sec:learning}, we can learn both the linear
projections of the features and the importance of each channel for the
retrieval task.  We demonstrate that such learning is feasible and can
significantly improve
performance.  We use a 50/50 split of the crime scene prints of the Israeli
dataset for training and testing, and determine hyperparameters settings using
10-fold cross-validation.  In the left panel of Fig.~\ref{fig:latent2trace} we
compare the performance of three different models with varying degrees of
learning.  The model with no learning is denoted as $[\mu_c,\sigma_c]$, with
learned per-channel weights is denoted as $[\mu_c,\sigma_c\cdot W_c]$, with
learned projections is denoted as CCA~$[\mu_c,\sigma_c]$, and with piece-wise
learned linear projections and per-channel weights is denoted as
CCA~$[\mu_c,\sigma_c\cdot W_c]$.  Our final model, CCA~$[\mu_c,\sigma_c\cdot
W_c]$~ft, jointly fine-tunes the linear projections and the per-channel weights
together.  The model with learned per-channel importance weights has $257$
parameters (a scalar for each channel and a single bias term), and was learned
using a support vector machine solver with a regularization value of $\alpha=100$.
The linear projections (CCA) were learned using \texttt{canoncorr}, MATLAB's
canonical correlation analysis function.  Our final model,
CCA~$[\mu_c,\sigma_c\cdot W_c]$~ft, was fine-tuned using gradient descent with 
an L2 regularization value of $\alpha=100$ on the per-channel importance weights
and $\beta=1$ on the linear projections. This full model has 131K parameters
($2\times 256^2$ projections, $256$ channel importance, and $1$ bias).

As seen in the left panel of Fig.~\ref{fig:latent2trace}, learning per-channel
importance weights, $[\mu_c,\sigma_c\cdot W_c]$, yields substantial 
improvements, outperforming $[\mu_c,\sigma_c]$ and CCA~$[\mu_c,\sigma_c]$ when
recall is less than 0.34.  When learning both importance weights and linear
projections, we see gains across all recall values as our Siamese network
significantly outperforms all other models. However, we observe only marginal
gains when fine-tuning the whole model. We expect this is due in part to the
small amount of training data which makes it difficult to optimize parameters
without overfitting.

We subsequently tested these same models (without any retraining) on the
FID-300 benchmark (shown in the right panel of Fig.~\ref{fig:latent2trace}).
In this, and in later experiments, we use cumulative match characteristic
(CMC) which plots the percentage of correct matches (recall) as a function of
the number of database items reviewed. This is more suitable for performance
evaluation than other information retrieval metrics such as precision-recall or
precision-at-k since there is only a single correct matching database item for
each query. CMC is easily interpreted in terms of the actually use-case scenario
(i.e., how much effort a forensic investigator must expend in verifying
putative matches to achieve a given level of recall).

On FID-300, we observe the same trend as on the Israeli dataset --- models with
more learned parameters perform better.  However, even without learning (\ie,
$[\mu_c,\sigma_c]$) MCNCC significantly outperforms using off-the-shelf
CNN features the previously published state-of-the-art approaches of
Kortylewski et
al.~\cite{kortylewski2014unsupervised,kortylewski2016probabilistic,kortylewski2017model}
The percentage of correct matches at top-1\% and top-5\% of the database image
reviewed for ACCV are 14.67 and 30.67, for BMVC16 are 21.67 and 47.00, for LoG16
are 59.67 and 73.33, for $[\mu_c,\sigma_c]$ are 72.67 and 82.33, and for
CCA~$[\mu_c,\sigma_c]$~ft are 79.67 and 86.33.
In Fig.~\ref{fig:basel_retrieval}, we visualize the top-10 retrieved test
impressions for a subset of crime scene query prints from FID-300.  These
results correspond to the CMC curves for $[\mu_c,\sigma_c]$ and
CCA~$[\mu_c,\sigma_c\cdot W_c]$ of the right panel of
Fig.~\ref{fig:latent2trace}.

\begin{figure*}[t]
\begin{center}
  \begin{tabular}{c@{\,}c@{\,}c@{\,}c@{\,}c@{\,}cc@{\,}c@{\,}c@{\,}c@{\,}c@{\,}c}
    & & \multicolumn{2}{c}{$[\cdot,\cdot]$} & \multicolumn{2}{c}{$[\mu_c,\sigma_c]$} &
    & & \multicolumn{2}{c}{$[\cdot,\cdot]$} & \multicolumn{2}{c}{$[\mu_c,\sigma_c]$} \\
    \includegraphics[trim=0 0 0 0,clip, width=12mm]{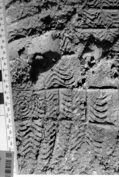} &
    \includegraphics[trim=0 0 0 0,clip, width=12mm]{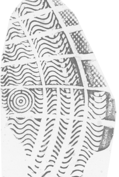} &
    \includegraphics[trim=0 0 0 0,clip, width=12mm]{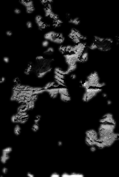} &
    \includegraphics[trim=0 0 0 0,clip, width=12mm]{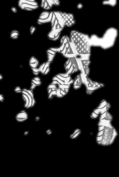} &
    \includegraphics[trim=0 0 0 0,clip, width=12mm]{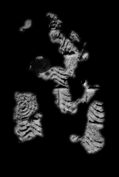} &
    \includegraphics[trim=0 0 0 0,clip, width=12mm]{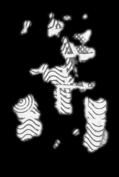} &
    \includegraphics[trim=0 0 0 0,clip, height=18mm]{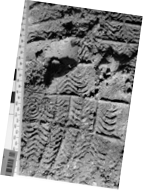} &
    \includegraphics[trim=0 0 0 0,clip, height=18mm]{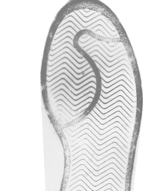} &
    \includegraphics[trim=0 0 0 0,clip, height=18mm]{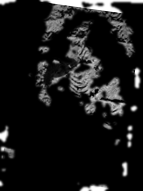} &
    \includegraphics[trim=0 0 0 0,clip, height=18mm]{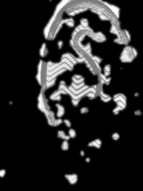} &
    \includegraphics[trim=0 0 0 0,clip, height=18mm]{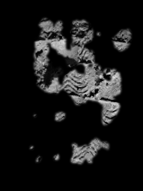} &
    \includegraphics[trim=0 0 0 0,clip, height=18mm]{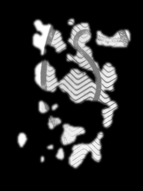} \\
    \includegraphics[trim=0 0 0 0,clip, width=12mm]{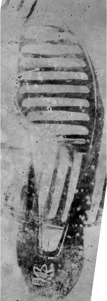} &
    \includegraphics[trim=0 0 0 0,clip, width=12mm]{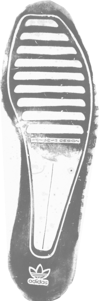} &
    \includegraphics[trim=0 0 0 0,clip, width=12mm]{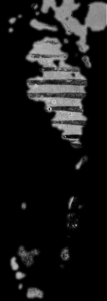} &
    \includegraphics[trim=0 0 0 0,clip, width=12mm]{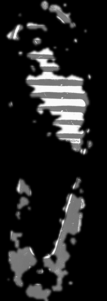} &
    \includegraphics[trim=0 0 0 0,clip, width=12mm]{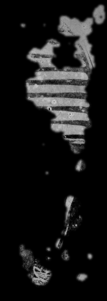} &
    \includegraphics[trim=0 0 0 0,clip, width=12mm]{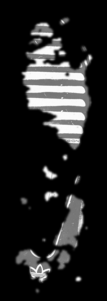} &
    \includegraphics[trim=2mm 0 7mm 0,clip, height=33mm]{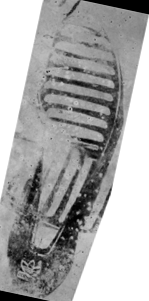} &
    \includegraphics[trim=8mm 0 1mm 0,clip, height=33mm]{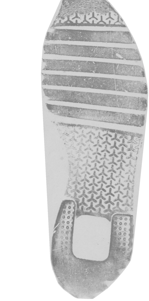} &
    \includegraphics[trim=8mm 0 1mm 0,clip, height=33mm]{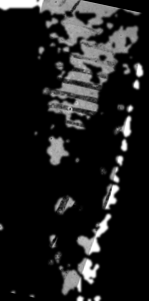} &
    \includegraphics[trim=8mm 0 1mm 0,clip, height=33mm]{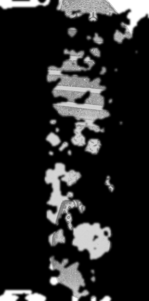} &
    \includegraphics[trim=8mm 0 1mm 0,clip, height=33mm]{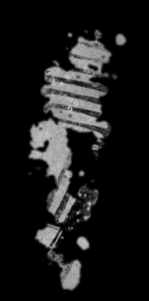} &
    \includegraphics[trim=8mm 0 1mm 0,clip, height=33mm]{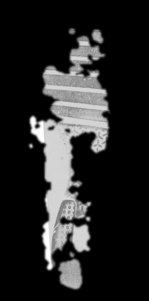} \\
    \includegraphics[trim=0 0 0 0,clip, width=12mm]{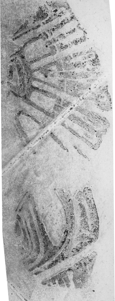} &
    \includegraphics[trim=0 0 0 0,clip, width=12mm]{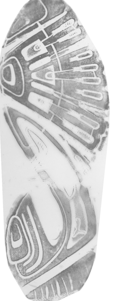} &
    \includegraphics[trim=0 0 0 0,clip, width=12mm]{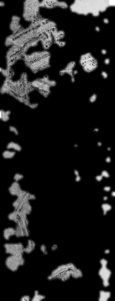} &
    \includegraphics[trim=0 0 0 0,clip, width=12mm]{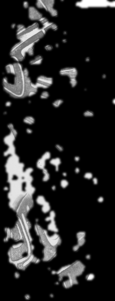} &
    \includegraphics[trim=0 0 0 0,clip, width=12mm]{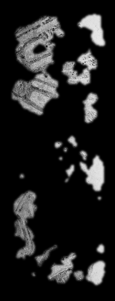} &
    \includegraphics[trim=0 0 0 0,clip, width=12mm]{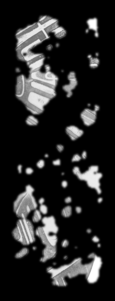} &
    \includegraphics[trim=0 0 12mm 0,clip, height=30mm]{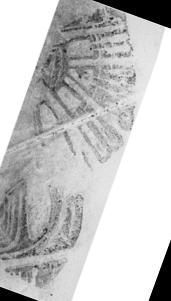} &
    \includegraphics[trim=8mm 0 4mm 0,clip, height=30mm]{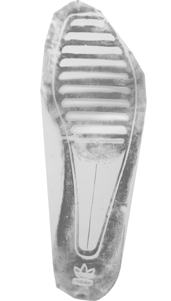} &
    \includegraphics[trim=8mm 0 4mm 0,clip, height=30mm]{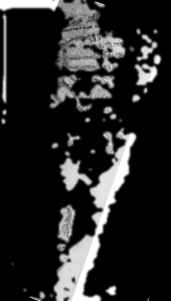} &
    \includegraphics[trim=8mm 0 4mm 0,clip, height=30mm]{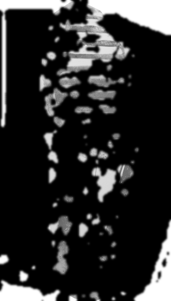} &
    \includegraphics[trim=8mm 0 4mm 0,clip, height=30mm]{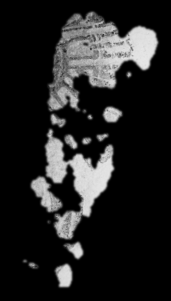} &
    \includegraphics[trim=8mm 0 4mm 0,clip, height=30mm]{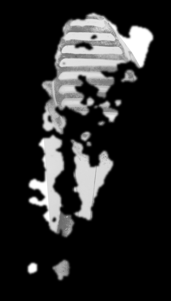} \\
  \end{tabular}
\end{center}\vspace{-2mm}
\caption{Visualizing image regions that have the greatest influence on positive
correlation between image pairs.  Each group of images shows, from left to
right, the original crime scene print and test impression being compared, the
image regions of the pair that have the greatest influence on positive
correlation score when using raw cross-correlation, and the image regions of
the pair that have greatest influence on positive MCNCC.  Each row shows the
same crime scene query aligned with a true matching impression (left) and with
a non-matching test impression (right). }
\label{fig:backprop_vis}
\end{figure*}

\begin{figure*}[t]
\begin{center}
\begin{minipage}{0.45\textwidth}
  \includegraphics[trim=0 0 0 0,clip, height=51mm]{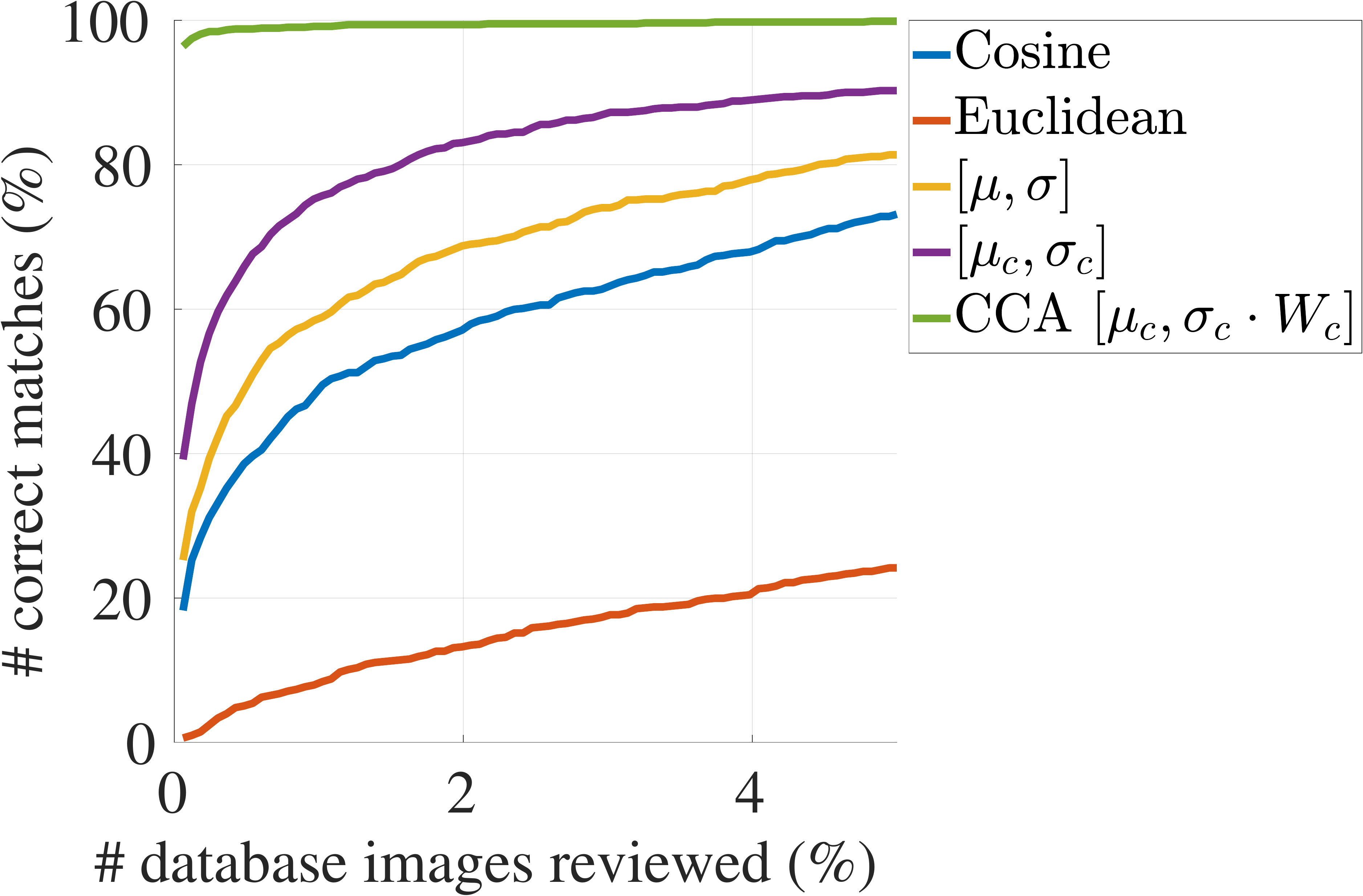}
\end{minipage}
\begin{minipage}{0.5\textwidth}
  \flushright
  \includegraphics[trim=0 0 0 0,clip, width=80mm]{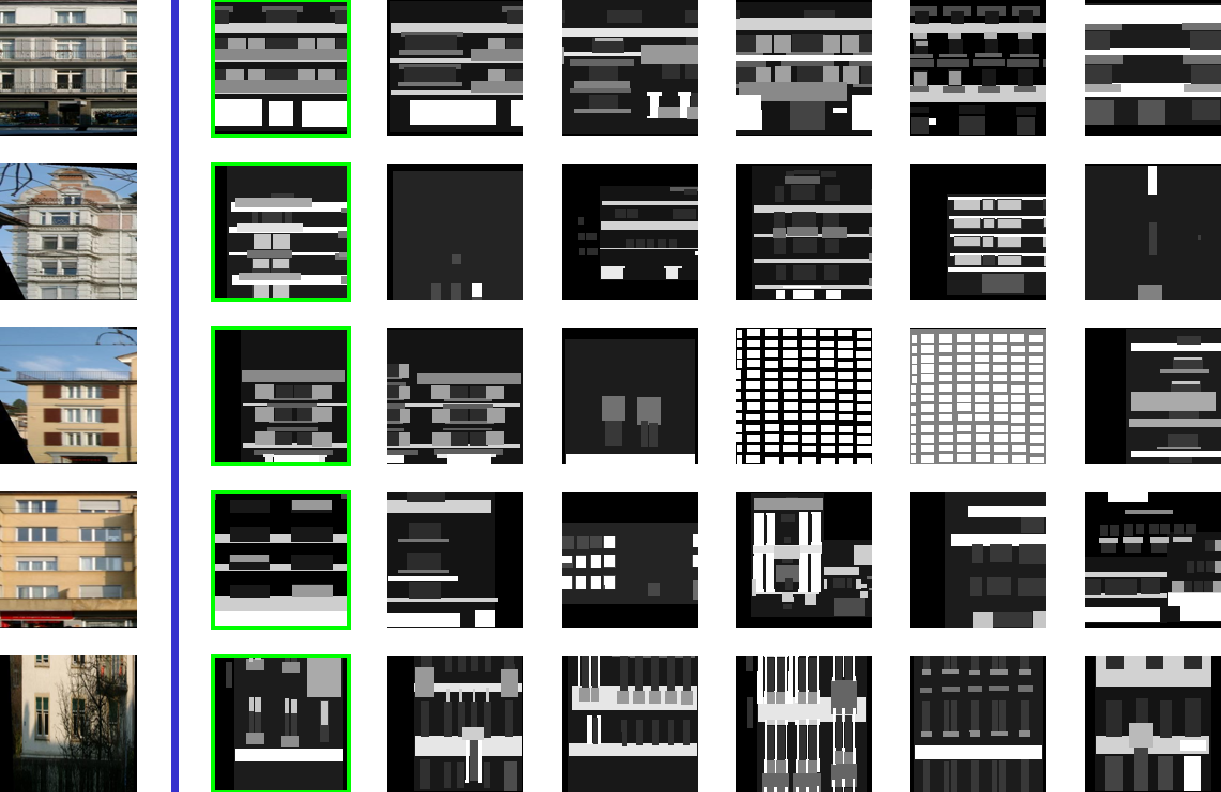}
\end{minipage}
\end{center}
\caption{Segmentation retrieval for building facades. The left panel compares
  MCNCC with learned linear projections and per-channel importance weights
  (denoted as CCA~$[\mu_c,\sigma_c\cdot W_c]$) and MCNCC with no learning
  (denoted as $[\mu_c,\sigma_c]$) to other baseline metrics: Cosine
  similarity, Euclidean distance, and NCC using across-channel local
  statistics (denoted as $[\mu,\sigma]$). The right panel shows example
  retrieval results for CCA~$[\mu_c,\sigma_c\cdot W_c]$. The left column shows
  the query facade image. Green boxes indicate the corresponding ground truth
  segmentation label.}
\label{fig:facade_retrieval}
\end{figure*}

\paragraph{Partial occlusion:}
To analyze the effect of partial occlusion on matching accuracy, we split the
set of crime scene query prints into subsets with varying amounts of occlusion.
For this we use the proxy of pixel area of the cropped crime scene print
compared to its corresponding test impression. The prints were then grouped
into 4 categories with roughly equal numbers of examples: ``Full size'' prints
are those whose pixel-area ratios fall between $[0.875, 1]$, ``$3/4$ size''
between $[0.625, 0.875)$, ``half size'' between $[0.375, 0.625)$, and ``$1/4$
size'' between $[0, 0.375)$.  In Table~\ref{tab:basel_ablation} we compare the
performance of models $[\mu_c,\sigma_c]$, CCA $[\mu_c,\sigma_c]$, and CCA
$[\mu_c,\sigma_c\cdot W_c]$. As expected, the correct match rate generally
increases for all models as the pixel area ratio increases and more
discriminative tread features are available, with the exception of ``full
size'' prints.  While ``full size'' query prints might be expected to include
more relevant features for matching, we have observed that in the benchmark
dataset they are often corrupted by additional ``noise'' in the form of
smearing or distortion of the print and marks left by overlapping impressions.

\begin{figure*}[t]
\begin{center}
\begin{minipage}{0.45\textwidth}
  \includegraphics[trim=0 0 0 0,clip, height=51mm]{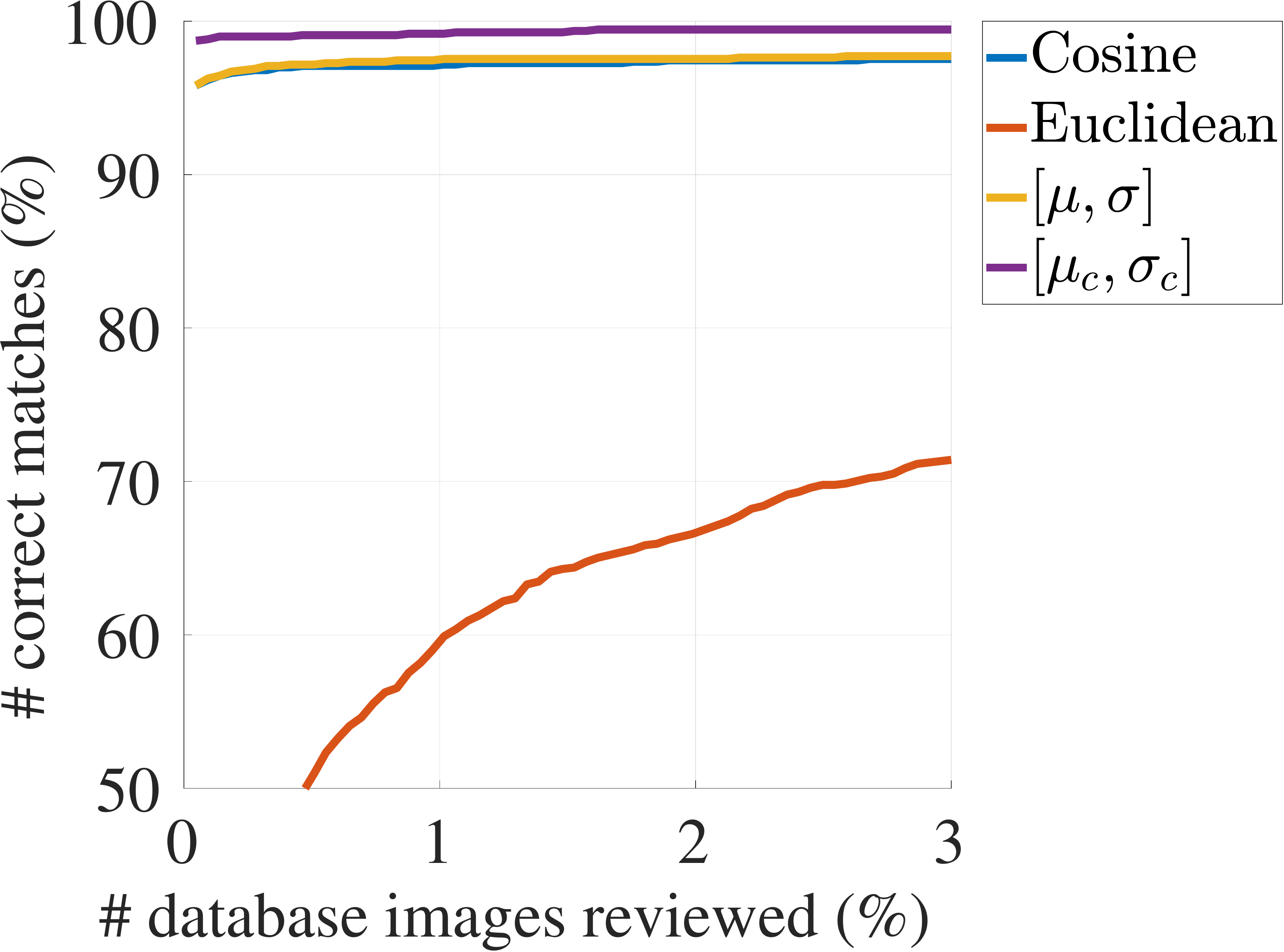}
\end{minipage}
\begin{minipage}{0.5\textwidth}
  \flushright
  \includegraphics[trim=0 0 0 0,clip, width=80mm]{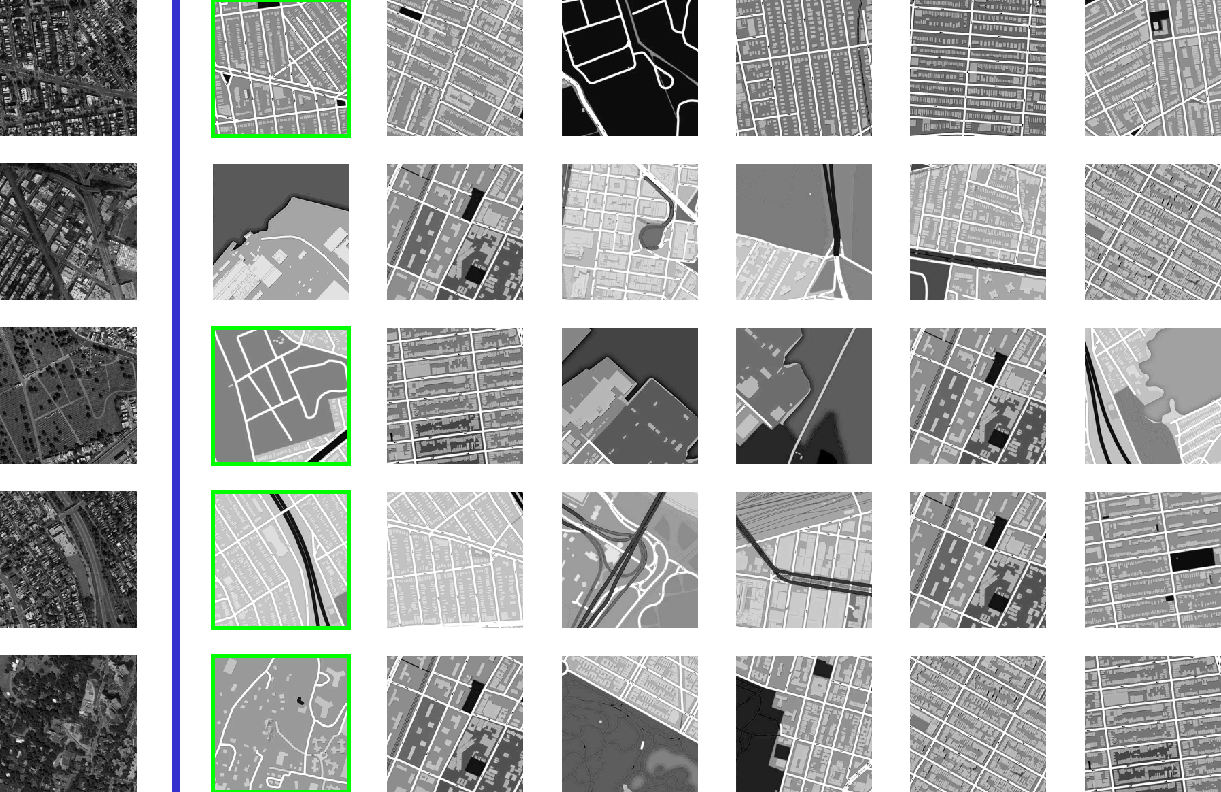}
\end{minipage}
\end{center}
\caption{Retrieval of maps from aerial imagery. The left panel compares
  MCNCC with no learning (denoted as $[\mu_c,\sigma_c]$) to other baseline
  metrics: Cosine similarity, Euclidean distance, and NCC using across-channel
  per-exemplar statistics (denoted as $[\mu,\sigma]$). The right panel shows
  retrieval results for $[\mu_c,\sigma_c]$. The left column shows the query
  aerial photo. Green boxes indicate the corresponding ground-truth map image.
  }
\label{fig:map_retrieval}
\end{figure*}

\paragraph{Background clutter:}
We also examined how performance was affected by the amount of irrelevant
background clutter in the crime scene print.  We use the ratio of the pixel
area of the cropped crime scene print over the pixel area of the original
crime scene print as a proxy for the amount of relevant information in a print.
Prints with a ratio closer to zero contain a lot of background, while prints
with a ratio closer to one contain little irrelevant information.  We selected
257 query prints with a large amount of background (ratio $\le 0.5$). 

When performing matching over these whole images we found that the percentage
of correct top-1\% matches dropped from 72.4\% to 15.2\% and top-10\% dropped
from 88.3\% to 33.5\%. This drop in performance is not surprising given that our
matching approach aims to answer the question of \emph{what} print is present,
rather than detecting \emph{where} a print appears in an image and was not
trained to reject background matches. We note that in practical investigative
applications, the quantity of footwear evidence is limited and a forensic
examiner would likely be willing to mark valid regions of query image, limiting
the effect of background clutter.

\paragraph{Visualizing image characteristics relevant to positive correlations:}
To get an intuitive understanding of what image features are utilized by MCNCC,
we visualize what image regions have a large influence the positive correlation
between paired crime scene prints and test impressions.  For a pair of images,
we backpropagate gradients to the image from each spatial bin in the feature
map which has a positive normalized correlation. We then produce a mask in the
image domain marking pixels whose gradient magnitudes are in the
top 20th percentile.  Fig.~\ref{fig:backprop_vis} compares this positive
relevance map for regular correlation (inner product of the raw features) and
normalized correlation (inner product of the standardized features).  We can
see that with normalized correlation, the image regions selected are similar
for both images despite the domain shift between the query and match.  In
contrast, the visualization for regular correlation shows much less coherence
across the pair of images and often attends to uninformative background edges
and blank regions.

\subsection{Segmentation Retrieval for Building Facades}
To further demonstrate the robustness of MCNCC for cross domain matching, we
consider the task of retrieving segmentation label maps which match for a given
building facade query image.  We use the CMP Facade
Database~\cite{tylecek2013spatial} which contains 606 images of facades from
different cities around the world and their corresponding semantic segmentation
labels. These labels can be viewed as a simplified ``cartoon image'' of the 
building facade by mapping each label to a distinct gray level.

In our experiments, we generate 1657 matching pair by resizing the original 606
images (base + extended dataset) to either $512 \times 1536$ or $1536 \times 512$
depending on their aspect ratio and crop out non-overlapping $512 \times 512$
patches.  We prune this set by removing 161 patches which contain more than
50\% background pixels to get our final dataset.  Examples from this dataset
can be seen in the right panel of Fig.~\ref{fig:facade_retrieval}. In order treat the 
segmentation label map as an image suitable for the pre-trained feature extractor,
we scale the segmentation labels to span the whole range of gray values (\ie,
from $[1-12]$ to $[0-255]$).

We compare MCNCC (denoted in the legend as $[\mu_c,\sigma_c]$) to three
baseline similarity metrics: Cosine, Euclidean distance, and
normalized cross-correlation using across-channel local statistics (denoted as
$[\mu,\sigma]$).  We can see in the left panel of Fig.~\ref{fig:facade_retrieval}
that MCNCC performs significantly better than the baselines. MCNCC returns the
true matching label map as the top scoring match in 39.2\% of queries. In
corresponding top match accuracy for normalized cross-correlation using
across-channel local statistics is 25.2\%, for Cosine similarity is 18.3\%, and
for Euclidean distance is 6.0\%.  When learning parameters with MCNCC (denoted
as CCA~$[\mu_c,\sigma_c\cdot W_c]$), using a 50/50 training-test split, we
see significantly better retrieval performance (96.4\% for reviewing one
database item).  The right panel of Fig.~\ref{fig:facade_retrieval} shows some
example retrieval results for this model.

\begin{table*}[t]
\begin{center}
\begin{tabular}{c|c|c|c|c|c|c}
  & & all prints & full size & $3/4$ size & half size & $1/4$ size \\
  \hline
  \# prints & & 300 & 88 & 78 & 71 & 63 \\
  \hline
  \multirow{3}{*}{Top-1\%} & $[\mu_c,\sigma_c]$ & 72.7 & 78.4 & 82.1 & 71.8 & 53.0 \\
  & CCA $[\mu_c,\sigma_c]$ & 76.8 & 83.0 & 85.9 & 73.2 & 60.3 \\
  & CCA $[\mu_c,\sigma_c\cdot W_c]$ & 79.0 & 84.1 & 85.9 & 78.9 & 63.5 \\
  \hline
  \multirow{3}{*}{Top-10\%} & $[\mu_c,\sigma_c]$ & 87.7 & 87.5 & 92.3 & 85.9 & 84.1 \\
  & CCA $[\mu_c,\sigma_c]$ & 88.7 & 93.2 & 91.0 & 87.3 & 81.0 \\
  & CCA $[\mu_c,\sigma_c\cdot W_c]$ & 89.3 & 93.2 & 91.0 & 91.6 & 79.4
\end{tabular}
\end{center}
\caption{Occlusion study on FID-300. The crime scene query prints are binned by
looking at the ratio of query pixel area to the pixel area of the corresponding
ground-truth test impression. Performance is measured as the percentage of
correct matches retrieved (higher is better).}
\label{tab:basel_ablation}
\end{table*}

\subsection{Retrieval of Maps from Aerial Imagery}
Finally, we evaluate matching performance on the problem of retrieving map data
corresponding to query aerial photos.  We use a dataset released by
Isola~\etal~\cite{isola2017image} that contains 2194 pairs of images scraped
from Google Maps. For simplicity in treating this as a retrieval task, we
excluded map tiles which consisted entirely of water.  Both aerial photos and
map images were converted from RGB to gray-scale prior to feature extraction
(see the right panel of Fig.~\ref{fig:map_retrieval} for examples).  We compare MCNCC to
three baseline similarity metrics: Cosine, Euclidean distance, and normalized
cross-correlation using across-channel local statistics (denoted as $[\mu,\sigma]$).

The results are shown in the left panel of Fig.~\ref{fig:map_retrieval}.  MCNCC
outperforms the baseline Cosine and Euclidean distance measures, but this time
performance of normalized cross-correlation using local per-exemplar statistics
averaged over all channels and Cosine similarity are nearly identical.  For
top-1 retrieval performance, MCNCC is correct 98.7\% of the time, normalized
cross-correlation using across-channel local statistics and Cosine similarity
are correct 95.8\%, and Euclidean distance is correct 28.6\% of the time when
retrieving only one item.  We show example retrieval results for MCNCC in the
right panel of Fig.~\ref{fig:map_retrieval}. We did not evaluate any learned
models in this experiment since the performance of baseline MCNCC left little
room for improvement.

\section{Conclusion}

In this work, we proposed an extension to normalized cross-correlation suitable
for CNN feature maps that performs normalization of feature responses on a
per-channel and per-exemplar basis. The benefits of performing per-exemplar
normalization can be explained in terms of spatially local whitening which
adapts to non-stationary statistics of the input.  Relative to other standard
feature normalization schemes (\eg, cosine similarity), per-channel
normalization accommodates variation in statistics of different feature
channels. 

Utilizing MCNCC in combination with CCA provides a highly effective building
block for constructing Siamese network models that can be trained in an
end-to-end discriminative learning framework.  Our experiments demonstrate that
even with very limited amounts of data, this framework achieves robust
cross-domain matching using generic feature extractors combined with piece-wise
training of simple linear feature-transform layers. This approach yields
state-of-the art performance for retrieval of shoe tread patterns matching
crime scene evidence.  We expect our findings here will be applicable to a wide
variety of single-shot and exemplar matching tasks using CNN features.

\begin{acknowledgements}
We thank Sarena Wiesner and Yaron Shor for providing access to their dataset. 
This work was partially funded by the Center for Statistics and Applications in
Forensic Evidence (CSAFE) through NIST Cooperative Agreement \#70NANB15H176.
\end{acknowledgements}

\bibliographystyle{spmpsci}      
\bibliography{refs.bib}

\begin{thebibliography}{10}
\providecommand{\url}[1]{{#1}}
\providecommand{\urlprefix}{URL }
\expandafter\ifx\csname urlstyle\endcsname\relax
  \providecommand{\doi}[1]{DOI~\discretionary{}{}{}#1}\else
  \providecommand{\doi}{DOI~\discretionary{}{}{}\begingroup
  \urlstyle{rm}\Url}\fi

\bibitem{bodziak1999footwear}
Bodziak, W.J.: Footwear impression evidence: detection, recovery and
  examination.
\newblock CRC Press (1999)

\bibitem{ChenetalSketch2Photo2009}
Chen, T., Cheng, M.M., Tan, P., Shamir, A., Hu, S.M.: Sketch2photo: Internet
  image montage.
\newblock In: ACM Transactions on Graphics (TOG), vol.~28, p. 124. ACM (2009)

\bibitem{CosteaBMVC2016}
Costea, D., Leordeanu, M.: Aerial image geolocalization from recognition and
  matching of roads and intersections.
\newblock arXiv preprint arXiv:1605.08323  (2016)

\bibitem{dardi2009texture}
Dardi, F., Cervelli, F., Carrato, S.: A texture based shoe retrieval system for
  shoe marks of real crime scenes.
\newblock Image Analysis and Processing--ICIAP 2009 pp. 384--393 (2009)

\bibitem{de2005automated}
De~Chazal, P., Flynn, J., Reilly, R.B.: Automated processing of shoeprint
  images based on the fourier transform for use in forensic science.
\newblock IEEE transactions on pattern analysis and machine intelligence
  \textbf{27}(3), 341--350 (2005)

\bibitem{DivechaSIGSPATIAL16}
Divecha, M., Newsam, S.: Large-scale geolocalization of overhead imagery.
\newblock In: Proceedings of the 24th ACM SIGSPATIAL International Conference
  on Advances in Geographic Information Systems, p.~32. ACM (2016)

\bibitem{fisher1995multi}
Fisher, R.B., Oliver, P.: Multi-variate cross-correlation and image matching.
\newblock In: Proc. British Machine Vision Conference (BMVC) (1995)

\bibitem{geiss1991multivariate}
Geiss, S., Einax, J., Danzer, K.: Multivariate correlation analysis and its
  application in environmental analysis.
\newblock Analytica chimica acta \textbf{242}, 5--9 (1991)

\bibitem{gueham2008automatic}
Gueham, M., Bouridane, A., Crookes, D.: Automatic recognition of partial
  shoeprints using a correlation filter classifier.
\newblock In: Machine Vision and Image Processing Conference, 2008. IMVIP'08.
  International, pp. 37--42. IEEE (2008)

\bibitem{hariharan2012discriminative}
Hariharan, B., Malik, J., Ramanan, D.: Discriminative decorrelation for
  clustering and classification.
\newblock Computer Vision--ECCV 2012 pp. 459--472 (2012)

\bibitem{isola2017image}
Isola, P., Zhu, J.Y., Zhou, T., Efros, A.A.: Image-to-image translation with
  conditional adversarial networks.
\newblock In: Proceedings of the IEEE Conference on Computer Vision and Pattern
  Recognition (2017)

\bibitem{KongSRF_BMVC_2017}
Kong, B., Supancic, J.S., Ramanan, D., Fowlkes, C.C.: Cross-domain forensic
  shoeprint matching.
\newblock In: British Machine Vision Conference (BMVC) (2017)

\bibitem{kortylewski2017model}
Kortylewski, A.: Model-based image analysis for forensic shoe print
  recognition.
\newblock Ph.D. thesis, University\_of\_Basel (2017)

\bibitem{kortylewski2014unsupervised}
Kortylewski, A., Albrecht, T., Vetter, T.: Unsupervised footwear impression
  analysis and retrieval from crime scene data.
\newblock In: Asian Conference on Computer Vision, pp. 644--658. Springer
  (2014)

\bibitem{kortylewski2016probabilistic}
Kortylewski, A., Vetter, T.: Probabilistic compositional active basis models
  for robust pattern recognition.
\newblock In: British Machine Vision Conference (2016)

\bibitem{lee2001advances}
Lee, H.C., Ramotowski, R., Gaensslen, R.: Advances in fingerprint technology.
\newblock CRC press (2001)

\bibitem{li2006one}
Li, F.F., Fergus, R., Perona, P.: One-shot learning of object categories.
\newblock IEEE Transactions on Pattern Analysis and Machine Intelligence
  \textbf{28}(4), 594--611 (2006)

\bibitem{malisiewicz2011ensemble}
Malisiewicz, T., Gupta, A., Efros, A.A.: Ensemble of exemplar-svms for object
  detection and beyond.
\newblock In: Computer Vision (ICCV), 2011 IEEE International Conference on,
  pp. 89--96. IEEE (2011)

\bibitem{MardiaKentBibby1980}
Mardia, K.V., Kent, J.T., Bibby, J.M.: Multivariate analysis (probability and
  mathematical statistics).
\newblock Academic Press London (1980)

\bibitem{martin1979multivariate}
Martin, N., Maes, H.: Multivariate analysis.
\newblock Academic press (1979)

\bibitem{parkhi2015deep}
Parkhi, O.M., Vedaldi, A., Zisserman, A.: Deep face recognition.
\newblock In: BMVC, vol.~1, p.~6 (2015)

\bibitem{patil2009rotation}
Patil, P.M., Kulkarni, J.V.: Rotation and intensity invariant shoeprint
  matching using gabor transform with application to forensic science.
\newblock Pattern Recognition \textbf{42}(7), 1308--1317 (2009)

\bibitem{pavlou2006automatic}
Pavlou, M., Allinson, N.: Automatic extraction and classification of footwear
  patterns.
\newblock Intelligent Data Engineering and Automated Learning--IDEAL 2006 pp.
  721--728 (2006)

\bibitem{popper1974multivariate}
Popper~Shaffer, J., Gillo, M.W.: A multivariate extension of the correlation
  ratio.
\newblock Educational and Psychological Measurement \textbf{34}(3), 521--524
  (1974)

\bibitem{tylecek2013spatial}
Radim~Tyle{\v c}ek, R.{\v S}.: Spatial pattern templates for recognition of
  objects with regular structure.
\newblock In: Proc. GCPR. Saarbrucken, Germany (2013)

\bibitem{Richetelli2017}
Richetelli, N., Lee, M.C., Lasky, C.A., Gump, M.E., Speir, J.A.: Classification
  of footwear outsole patterns using fourier transform and local interest
  points.
\newblock Forensic science international \textbf{275}, 102--109 (2017)

\bibitem{RusselAlignment2011}
Russell, B.C., Sivic, J., Ponce, J., Dessales, H.: Automatic alignment of
  paintings and photographs depicting a 3d scene.
\newblock In: Computer Vision Workshops (ICCV Workshops), 2011 IEEE
  International Conference on, pp. 545--552. IEEE (2011)

\bibitem{SenletICPR2014}
Senlet, T., El-Gaaly, T., Elgammal, A.: Hierarchical semantic hashing: Visual
  localization from buildings on maps.
\newblock In: Pattern Recognition (ICPR), 2014 22nd International Conference
  on, pp. 2990--2995. IEEE (2014)

\bibitem{sharif2014cnn}
Sharif~Razavian, A., Azizpour, H., Sullivan, J., Carlsson, S.: Cnn features
  off-the-shelf: an astounding baseline for recognition.
\newblock In: Proceedings of the IEEE Conference on Computer Vision and Pattern
  Recognition Workshops, pp. 806--813 (2014)

\bibitem{ShrivastavaCrossDomain2011}
Shrivastava, A., Malisiewicz, T., Gupta, A., Efros, A.A.: Data-driven visual
  similarity for cross-domain image matching.
\newblock ACM Transactions on Graphics (ToG) \textbf{30}(6), 154 (2011)

\bibitem{tang2010footwear}
Tang, Y., Srihari, S.N., Kasiviswanathan, H., Corso, J.J.: Footwear print
  retrieval system for real crime scene marks.
\newblock In: International Workshop on Computational Forensics, pp. 88--100.
  Springer (2010)

\bibitem{wei2014alignment}
Wei, C.H., Gwo, C.Y.: Alignment of core point for shoeprint analysis and
  retrieval.
\newblock In: Information Science, Electronics and Electrical Engineering
  (ISEEE), 2014 International Conference on, vol.~2, pp. 1069--1072. IEEE
  (2014)

\bibitem{xiao2016learning}
Xiao, T., Li, H., Ouyang, W., Wang, X.: Learning deep feature representations
  with domain guided dropout for person re-identification.
\newblock In: Proceedings of the IEEE Conference on Computer Vision and Pattern
  Recognition, pp. 1249--1258 (2016)

\bibitem{yekutieli2012expert}
Yekutieli, Y., Shor, Y., Wiesner, S., Tsach, T.: Expert assisting computerized
  system for evaluating the degree of certainty in 2d shoeprints.
\newblock Tech. rep., Technical Report, TP-3211, National Institute of Justice
  (2012)

\bibitem{zagoruyko2015learning}
Zagoruyko, S., Komodakis, N.: Learning to compare image patches via
  convolutional neural networks.
\newblock In: Proceedings of the IEEE Conference on Computer Vision and Pattern
  Recognition, pp. 4353--4361 (2015)

\bibitem{zbontar2015computing}
Zbontar, J., LeCun, Y.: Computing the stereo matching cost with a convolutional
  neural network.
\newblock In: Proceedings of the IEEE Conference on Computer Vision and Pattern
  Recognition, pp. 1592--1599 (2015)

\bibitem{zhang2005automatic}
Zhang, L., Allinson, N.: Automatic shoeprint retrieval system for use in
  forensic investigations.
\newblock In: UK Workshop On Computational Intelligence (2005)

\end{thebibliography}

\end{document}